\documentclass[iicol]{sn-jnl}% Math and Physical Sciences Numbered Reference Style 
\usepackage{graphicx}%
\usepackage{multirow}%
\usepackage{amsmath,amssymb,amsfonts}%
\usepackage{amsthm}%
\usepackage{array}
\usepackage{mathrsfs}%
\usepackage[title]{appendix}%
\usepackage{textcomp}%
\usepackage{manyfoot}%
\usepackage{booktabs}%
\usepackage{multicol}
\usepackage{multirow}
\usepackage{algorithm}%
\usepackage{algorithmicx}%
\usepackage{algpseudocode}%
\usepackage{listings}%

\usepackage{float}
\usepackage[caption=false,font=normalsize,labelfont=sf,textfont=sf]{subfig}
\usepackage{stfloats}
\usepackage{url}
\usepackage{verbatim}

\usepackage{times}
\usepackage{epsfig}

\usepackage{bm}
\usepackage{microtype}
\usepackage{tablefootnote}

\usepackage[table]{xcolor}

\usepackage[capitalize]{cleveref}
\crefname{section}{Sec.}{Secs.}
\Crefname{section}{Section}{Sections}
\Crefname{table}{Table}{Tables}
\crefname{table}{Tab.}{Tabs.}

\def\Rcolor{\color[rgb]{0,0,0}}
\def\revise#1{\textcolor[rgb]{0,0,0}{#1}}

\usepackage{xspace}

\makeatletter
\DeclareRobustCommand\onedot{\futurelet\@let@token\@onedot}
\def\@onedot{\ifx\@let@token.\else.\null\fi\xspace}

\def\eg{\emph{e.g}\onedot} 
\def\ie{\emph{i.e}\onedot}

\makeatother

\begin{document}

\title[Article Title]{DreamArtist: Controllable One-Shot Text-to-Image Generation via Positive-Negative Adapter}


\author[1,2]{\fnm{Ziyi} \sur{Dong}}
\author[1,2]{\fnm{Pengxu} \sur{Wei}}
\author*[1,2]{\fnm{Liang} \sur{Lin}}

\affil[1]{\orgdiv{School of Computer Science and Technology}, \orgname{Sun Yat-sen Unviersity}, \orgaddress{\city{Guangzhou}, \country{China}}}

\affil[2]{\orgname{Peng Cheng Laboratory}, \orgaddress{\city{Shenzhen}, \country{China}}}

% \author*[1,2]{\fnm{First} \sur{Author}}\email{iauthor@gmail.com}



% \affil*[1]{\orgdiv{Department}, \orgname{Organization}, \orgaddress{\street{Street}, \city{City}, \postcode{100190}, \state{State}, \country{Country}}}

% \affil[2]{\orgdiv{Department}, \orgname{Organization}, \orgaddress{\street{Street}, \city{City}, \postcode{10587}, \state{State}, \country{Country}}}

% \affil[3]{\orgdiv{Department}, \orgname{Organization}, \orgaddress{\street{Street}, \city{City}, \postcode{610101}, \state{State}, \country{Country}}}


\abstract{

State-of-the-arts text-to-image generation models such as Imagen \cite{ImgGen} and Stable Diffusion Model \cite{LDM} have succeed remarkable progresses in synthesizing high-quality, feature-rich images with high resolution guided by human text prompts. Since certain characteristics of image content \emph{e.g.}, very specific object entities or styles, are very hard to be accurately described by text, some example-based image generation approaches have been proposed, \emph{i.e.} generating new concepts based on absorbing the salient features of a few input references. Despite of acknowledged successes, these methods have struggled on accurately capturing the reference examples' characteristics while keeping diverse and high-quality image generation, particularly in the one-shot scenario (\emph{i.e.} given only one reference). To tackle this problem, we propose a simple yet effective framework, namely DreamArtist, which adopts a novel positive-negative prompt-tuning learning strategy on the pre-trained diffusion model, and it has shown to well handle the trade-off between the accurate controllability and fidelity of image generation with only one reference example. Specifically, our proposed framework incorporates both positive and negative embeddings or adapters and optimizes them in a joint manner. The positive part aggressively captures the salient characteristics of the reference image to drive diversified generation and the negative part rectifies inadequacies from the positive part. We have conducted extensive experiments and evaluated the proposed method from image similarity (fidelity) and diversity, generation controllability, and style cloning. And our DreamArtist has achieved a superior generation performance over existing methods. Besides, our additional evaluation on extended tasks, including concept compositions and prompt-guided image editing, demonstrates its effectiveness for more applications.

\textit{DreamArtist project page: \url{https://www.sysu-hcp.net/projects/dreamartist/index.html}
}}




\maketitle

\section{Introduction}

\rightline{\emph{``Imagination will take you everywhere."}}
\rightline{\emph{---Albert Einstein.}}
%--\small{\sffamily {Albert Einstein}. } 

\vspace{6pt}
With productive imaginations and fantastic inspirations, everyone can be an artist, creating and being creative, which is the goal of the recently rising visual content generation research~\cite{LDM, ImgGen, Dalle2}. Thanks to the exponential evolution of generative models~\cite{sty3, DDIM, Glow, BigGAN, Palette, DiffGans, UNet, ScoreDiff}, we have witnessed the rapid progress of GAN and diffusion models on Text-to-Image synthesis~\cite{GLIDE, RiFeGAN, ZeroShotGen, ctrlGAN, MirrorGAN, DFGAN, TIGAN, VQGAN-CLIP, PPGAN}. Even more inspiring, given only texts with classifier~\cite{guid_cls} or classifier-free~\cite{guid_cls_free} guidance, large-scale text-to-image models~\cite{Dalle2, CogView, ImgGen, Parti, ERNIE}, such as LDM~\cite{LDM}, enable the synthesis of high-resolution images with rich details and various characteristics, fulfilling our diverse \emph{personalized} requirements. Despite yielding impressive images, these models require numerous words to depict a desirable complex image. Furthermore, they may struggle with words describing new concepts, styles, or object entities for image generation.
%%%%%

\begin{figure}[t]
    \centering
    \includegraphics[width=0.48\textwidth]{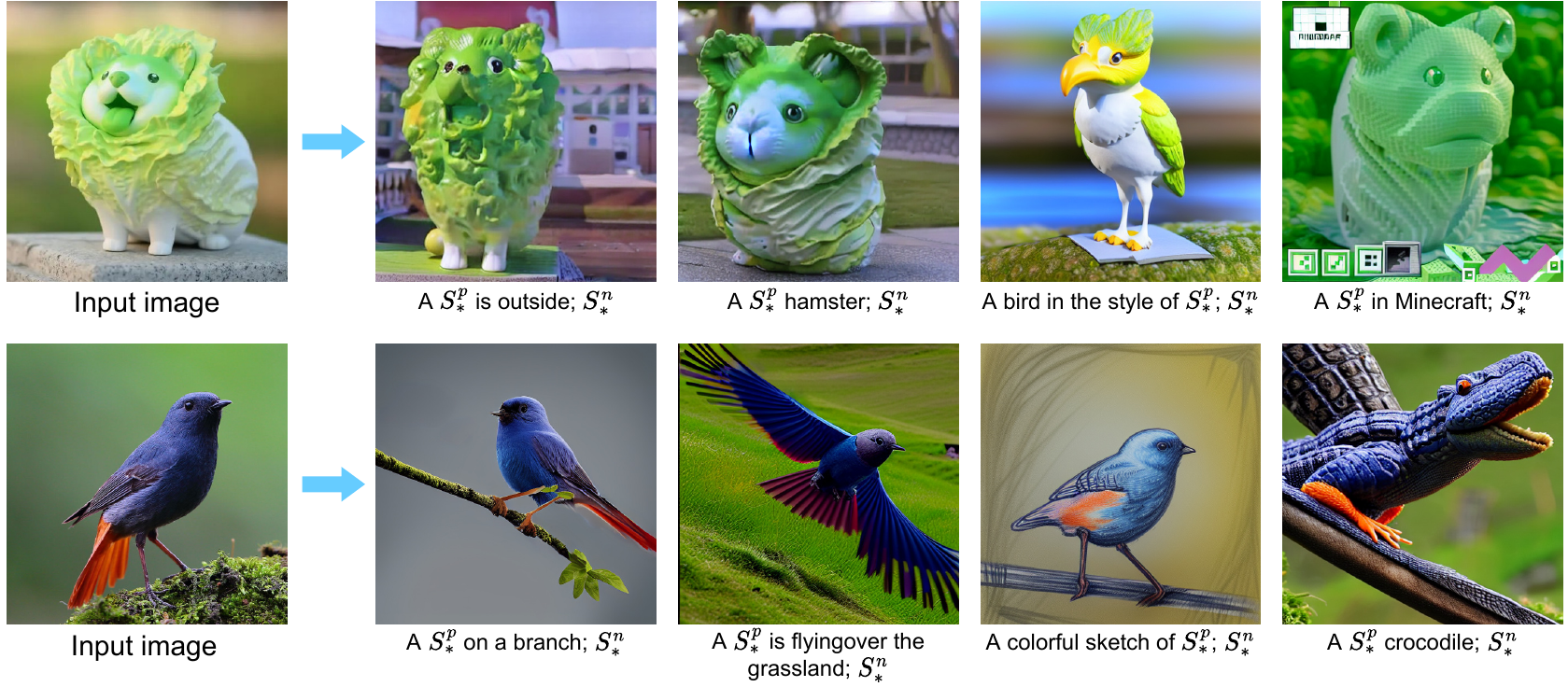}
    \vspace{-10pt}
    \caption{
    DreamArtist excels in learning to generate relevant, high-quality, diverse, and highly controllable images from only one reference image. Furthermore, it has the ability to incorporate certain abstract features from the reference image to create novel visual compositions.
    }
    %\vspace{-10pt}
    \label{fig:DA_head}
\end{figure}

\begin{figure*}[t]
    \centering
    \includegraphics[width=1\textwidth]{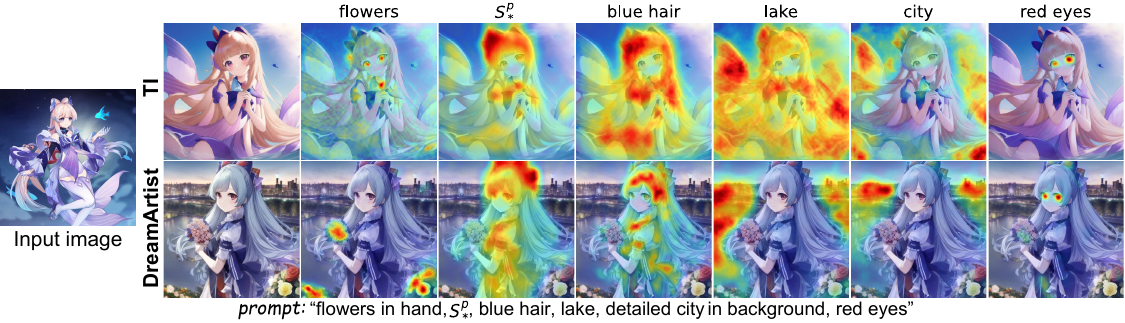}
    \vspace{-10pt}
    \caption{
    Attention maps (DAAM~\cite{DAAM}) of each token of the given text guidance in one-shot text-to-image generation. As prompt-tuning based methods, TI~\cite{TI} and our DreamArtist learn pseudo-word $S_*$ from the input image. 
    %Compared with TI, our DreamArtist can accurately render the visual content specified by the given texts.
    In comparison to TI, DreamArtist demonstrates a superior ability to accurately render the visual content specified by the given text descriptions.
    }
    \vspace{-10pt}
    \label{fig:comparson_attention}
\end{figure*}

% %\textbf{Related work.} 
% To alleviate this problem, very few attempts have been made. Two exemplars are Textual Inversion (TI)~\cite{TI} and DreamBooth~\cite{DreamBooth}. 
% They teach a pre-trained large-scale text-to-image model a new concept from 3-5 images and phrase it with a new pseudo-word. Then this new word can be quoted in the text-to-image generation process. 
% %
% Specifically, 
% DreamBooth~\cite{DreamBooth} employs the \emph{fine-tuning} strategy on a pre-trained model and learns to bind a unique class-specific pseudo-word with that new concept.
% %
% TI~\cite{TI} learns an embedding as pseudo-word $S_*$ to represent concepts in input images by \emph{prompt-tuning}. 

\begin{figure}[t]
    \centering
    \includegraphics[width=0.48\textwidth]{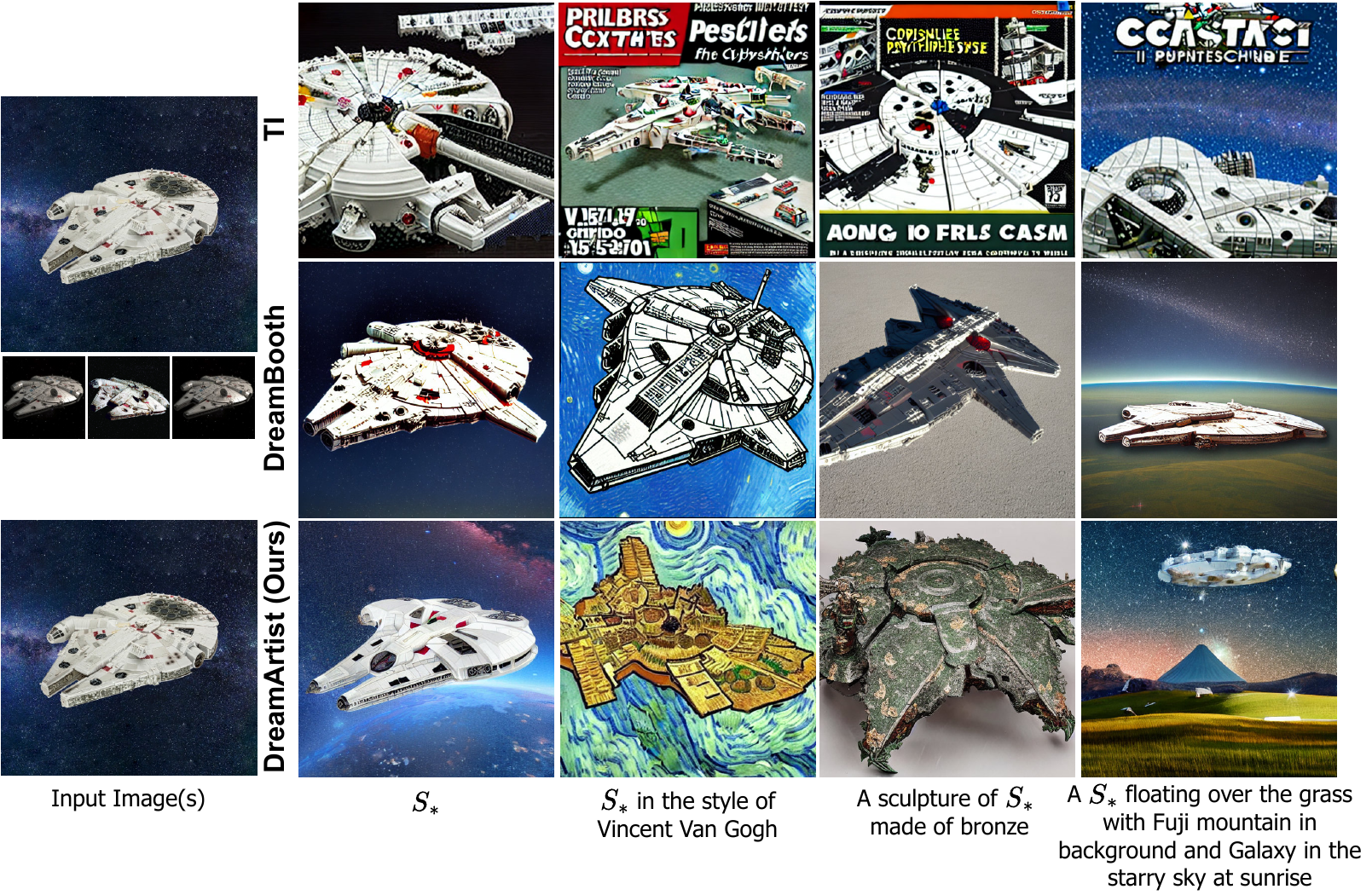}
    %\vspace{-18pt}
    \caption{
    Given a reference set with 3-5 images, TI~\cite{TI} and DreamBooth~\cite{DreamBooth} generate different images under different text guidance via learning a pseudo-word $S_*$. For comparison, given only \textbf{one} reference image, our DreamArtist can generate diverse images in different contexts, styles, materials, or others that are specified by given texts, presenting a better controllability.
    }
    \vspace{-10pt}
    \label{fig:comparison_TI_DB}
\end{figure}

%%%%%
To alleviate this problem, few attempts have been made, with Textual Inversion (TI)~\cite{TI} and DreamBooth~\cite{DreamBooth} being two notable examples. 
These methods aim to teach a pre-trained large-scale text-to-image model a new concept from a limited set of 3-5 images and associate it with a new pseudo-word. This newly learned pseudo-word can then be incorporated into the text-to-image generation process. 
Specifically, DreamBooth~\cite{DreamBooth} employs a \emph{fine-tuning} strategy on a pre-trained model, learning to bind a unique class-specific pseudo-word with the new concept. In contrast, TI~\cite{TI} learns an embedding as a pseudo-word $S_*$ to represent concepts in input images through \emph{prompt-tuning}.
Despite the significant potential for image generation offered by these methods, they also show several limitations. DreamBooth necessitates fine-tuning the pre-trained model, i.e., optimizing a vast number of parameters (see Table \ref{tab:cmp}), using only a few reference images. This approach often results in generated images that are monotonous and lack diversity, as illustrated in Figures \ref{fig:comparison_TI_DB} and \ref{fig:comparison_OneShotSetting}.
On the other hand, TI employs an energy-efficient prompt-tuning technique by optimizing only a limited number of parameters. Nevertheless, a notable drawback of this method is its usually failure to faithfully generate image details specified by prompt keywords. As depicted in Figure \ref{fig:comparson_attention}, crucial elements such as "flowers," "lake," and "city" are absent in the images generated by TI. Even with an increase in the number of reference images (from one-shot to 3-5 instances), this issue persists, as evidenced in Figure \ref{fig:comparison_TI_DB}. The primary reason for these shortcomings lies in the absence of a more effective learning strategy to enhance generation controllability during prompt-tuning with a minimal number of reference images.
Furthermore, incorporating multiple reference images may introduce unexpected ambiguities during the generation process, while gathering a collection of instances for a specific object would require additional effort.
Therefore, it is desirable that the method can generate image content to be precisely aligned with the original texts, while emphasizing the key elements extracted from the reference images. 

%%%%%

In this work, we introduce \emph{DreamArtist}, a simple yet effective approach for one-shot text-to-image generation. \emph{DreamArtist} is designed to learn a new concept from a single reference image (not a reference set), utilizing a pre-trained diffusion model. This method incorporates a \emph{positive-negative adapter} learning strategy, which adeptly balances the preservation of the reference's specific characteristics with text-guided generation controllability. 
Contrary to traditional methods such as prompt-tuning or fine-tuning (including adapters like LoRA~\cite{LoRA}), which learn an embedding ($S_*$) or a model ($\epsilon_\theta$) through positive guidance, \emph{DreamArtist} acquires both positive and negative embeddings ($S^p_*$ and $S^n_*$) as well as adapters ($\phi^p$ and $\phi^n$). These embeddings and adapters are based on the pre-trained text encoder $\mathcal{B}$ and denoising U-Net $\epsilon_\theta$.
Specifically, the positive components ($S^p_*$ and $\phi^p$) aggressively capture the characteristics of the reference image, thereby promoting diverse generation. Conversely, the negative components ($S^n_*$ and $\phi^n$) introspect in a self-induced manner to rectify the limitations inherent in the positive components. The inclusion of negative components introduces corrective information and facilitate rectifying the new concept from a differentiated (negative) aspect.
The integration of these negative components relaxes the optimization objective, moving beyond a rigid adherence to the input reference. This shift facilitates the generation of highly diverse images and significantly enhances controllability, allowing the seamless integration of newly learned concepts with the original visual content.
Importantly, since $S^n_*$ serves to rectifies $S^p_*$ specifically, the disturb on other textual features is minimal. We have conducted extensive experiments on the natural image dataset LAION-2B~\cite{LAION} and the anime dataset Danbooru~\cite{danbooru2021}. The results demonstrate that our \emph{DreamArtist} method achieves a substantial improvement over existing techniques.   
%%%%%

\begin{table}[t]
%\vspace{-3pt}
\centering
\footnotesize
\renewcommand\arraystretch{1}
\setlength\tabcolsep{3pt}
%\renewcommand\arraystretch{1.5}
%\resizebox{\columnwidth}{!}{

\caption{Comparison with current state-of-the-art methods.}

\begin{tabular}{|c||c|c|c|}
\hline
Method                  & TI~\cite{TI}                        & DreamBooth~\cite{DreamBooth}         & Ours           \\ \hline
Given image number          & 3-5                       & 3-5                & 1              \\ \hline
Parameters              & 2K                        & 983M            & 5K             \\ \hline
Image quality           & fair, mosaic & vivid & vivid         \\ \hline
%details                 & poor                      & awsome             & awsome         \\ \hline
%feature controllability & struggling                & partial            & completely     \\ \hline
Diversity               & limited                      & limited          & highly diverse \\ \hline
%style learning ability  & poor                      & avarage            & awsome         \\ \hline
%destroy existing features & no                        & high               & no             \\ \hline
Controllability & limited                & limited            & high     \\ \hline
\end{tabular}
%}

\label{tab:cmp}
\end{table}

Overall, our contributions are summarized as follows: 

\begin{figure*}[!t]
    \centering
    \includegraphics[width=1\linewidth]{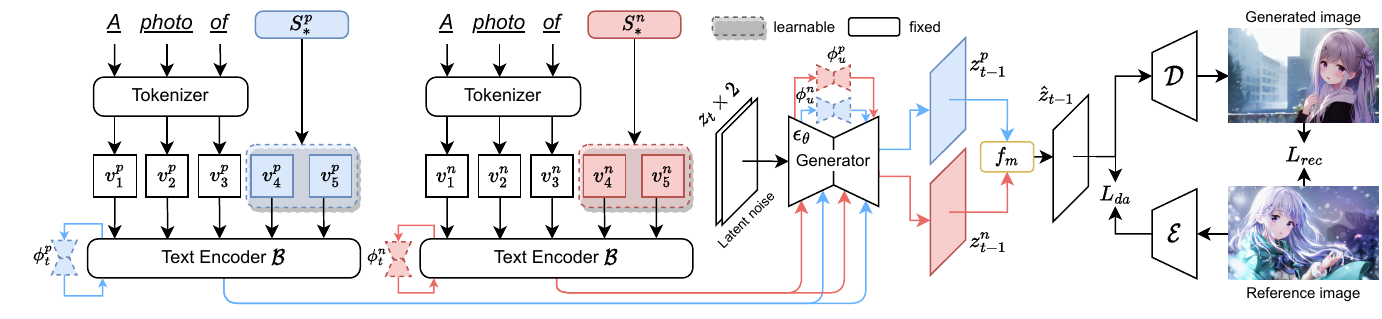}
    \vspace{-10pt}
    \caption{Framework of our DreamArtist.
    Only the embeddings corresponding to positive and negative pseudo-words ($S^p_*$ and $S^n_*$) need to be learned, and the rest of the parameters are fixed. $f_m=z^n + \gamma(z^p-z^n)$ is the classifier-free guidance of $z^p$ and $z^n$.
    }
    \vspace{-1ex}
    \label{fig:framework}
\end{figure*}

%First, we propose a novel one-shot text-to-image approach, DreamArtist, which is capable of producing what you dream about in painting based on one reference image.
\revise{First, we propose a novel one-shot text-to-image approach, DreamArtist, which enables users to express their creativity in painting based on one reference image.}

Second, we propose a general positive-negative adapter,  facilitating diverse and highly controllable image generation from the positive guidance with the complementary negative rectification. 

\textls[-10]{At last, we have conducted extensive qualitative and quantitative experiments on both natural and anime data, demonstrating that our method substantially outperforms existing methods in content quality, style similarity, diversity, and detail quality. 
%Our approach can render realistic images with learned features even combined with complex additional texts.
}

\section{Related Work}

\subsection{Diffusion-Based Generative Models}
In recent times, diffusion-based image generation models have achieved remarkable success. The first approach, known as the DDPM, proposed a method to iteratively denoise a noisy image and generate the image progressively. Compared to GAN-based methods, DDPM is more stable in training and generates more diverse images. However, the generation process of DDPM requires thousands of iterations, making it impractical to reality scenarios. To address this limitation, several algorithms such as DDIM, K-Diffusion, and DPM solver have been developed to accelerate the sampling and denoising process of DDPM. 

Another aspect, \cite{guid_cls} introduced a classifier guidance mechanism, which 
utilize the gradient of the log likelihood of an auxiliary classifier model $p_{\theta}(z_t|c)$ to guide the denoising process with condition $c$:
\begin{equation}
\tilde{\epsilon}_\theta\left(z_t, c\right)=\epsilon_\theta\left(z_t, c\right)-\gamma \sigma_t \nabla_{z_t} \log p_\theta\left(c \mid z_t\right).
\end{equation}
It enhances the conditional controllability of the diffusion model's generation and, for the first time, surpasses the performance of GAN-based methods. 
Furthermore, \cite{guid_cls_free} proposed the classifier-free guidance (CFG) method, enabling the diffusion model to simultaneously learn both conditional ($p_{\theta}(z_t|c)$) and unconditional ($p_{\theta}(z_t)=p_{\theta}(z_t|\emptyset)$) image generation. 
By constructing an implicit classifier $p_{\theta}(z_t|c) \propto \frac{p_{\theta}(z_t|c)}{p_{\theta}(z_t)}$, the performance of the model can be significantly enhanced.
%By leveraging Bayesian formula for guidance, this approach significantly improves the quality of generated images.

\subsection{Text-to-Image Synthesis}
With the exponential evolution of generative models, the focus of research on Text-to-Image synthesis has gradually shifted from GAN to Diffusion~\cite{GLIDE, RiFeGAN, ZeroShotGen, ctrlGAN, MirrorGAN, DFGAN}. Some large-scale text-to-image models~\cite{Dalle2, CogView, Parti, LDM} have made highly accurate and fine-grained controllable semantic generation. The recently proposed stable diffusion~\cite{LDM} unprecedented making high-resolution and high-quality large-scale text-to-image models become reality. These Diffusion models usually employ classifier guidance~\cite{guid_cls} or classifier-free~\cite{guid_cls_free} guidance to generate images with text guiding. 
\revise{\cite{Cnet, T2IA} attempts to add spatial conditioning controls pretrained text-to-image diffusion models, while \cite{IPA} employs both image and text as prompts to control the diffusion models with an additional cross attention.}
However, it is difficult for these methods to generate images following user-given patterns. And complex descriptions are required to generate high quality images.

\subsection{Few-Shot Text-to-Image Generation}
The diffusion-based Text-to-Image generation models have shown the ability to accurately and controllably generate high-quality images based on natural language descriptions. However, these models are not applicable to entirely novel concepts provided by users.
Therefore, some attempts try to tune diffusion model with a small image set, enabling the model to guide the denoising process toward those specific features.
\cite{TI} proposes the TI method trying to find a pseudo-words in the text vector space to represent the personalized object via prompt-tuning.
\cite{DreamBooth}, instead, proposes DreamBooth attempt to fine-tuning the entire model with a small image set under the premise of known personalized object categories.
\cite{LoRA} employs a reparameterizable Adapter to fine-tune a small subset of 
model parameters, which can mitigate overfitting to some extent.
\cite{customdiffusion} combining the advantages of DreamBooth and TI, trains only the $k$ and $v$ layer of the cross attention and introduces prompt-tuning together.

While these methods are able to learn the features of an object from few images, these methods suffer from overfitting and poor controllability. These methods still require 3-5 images, while our method requires only 1 image to generate highly controllable personalized features, which can be easily used with complex descriptions.

{\Rcolor
\section{Preliminary}

\subsection{Latent Diffusion Model}
With the remarkable capacity of image generation, Latent Diffusion Model (LDM)~\cite{LDM} is utilized as the base model. Different from DDPM~\cite{DDPM, DDIM} that performs denoising operations in the image space, LDM conducts this in the feature space. This readily facilitates the diffusion operations in the feature space. 
Formally, firstly, an input image $x$ is encoded into the feature space by an AutoEncoder (with an encoder $\mathcal{E}$ and a decoder $\mathcal{D}$, \ie,  $z=\mathcal{E}(x), \hat{x}=\mathcal{D}(z)$), pre-trained with a large number of images~\cite{LDM}. 
$t$ indicates the time step, and $z_t$ is the diffusion feature map of $z$ at the $t$-th step. 
At the $t$-th denoising step, a Denoising U-Net $\epsilon_{\theta}$ equipped with transformer blocks is used to perform denoising on the feature map $z_{t-1}=\epsilon_{\theta}(z_t, t)$. 
For text-guided conditional image generation, LDM utilizes a pre-trained text encoder $\mathcal{B}$ for given texts $S$ and has its text feature $y=\mathcal{B}(S)$. It employs the cross-attention mechanism with the image feature as query and two transformations of the text feature as key and value.  
Its training loss is formulated as
%
% \vspace{-8pt}
\begin{equation}
\mathcal {L}_{L D M}=\mathbb{E}_{\mathcal{E}(x), y, \epsilon \sim \mathcal{N}(0,1), t}\left[\left\|\epsilon-\epsilon_\theta\left(z_t, t, \tau_\theta(y)\right)\right\|_2^2\right]
\end{equation}
where %$t$ represents the time step, $z_t$ is the diffusion feature map of $z$ at the $t$-th step, and 
$\epsilon$ is the unscaled noise~\cite{LDM} and $\| \cdot \|^2_2$ is the $\ell 2$ loss. In this training phase, AutoEncoder is fixed and only $\epsilon_{\theta}$ is learnable.%involved in the optimization. 
%\pengxuc{no defination of $\tau$ and $theta$}

\begin{figure}[t!]
    \centering
    \includegraphics[width=0.48\textwidth]{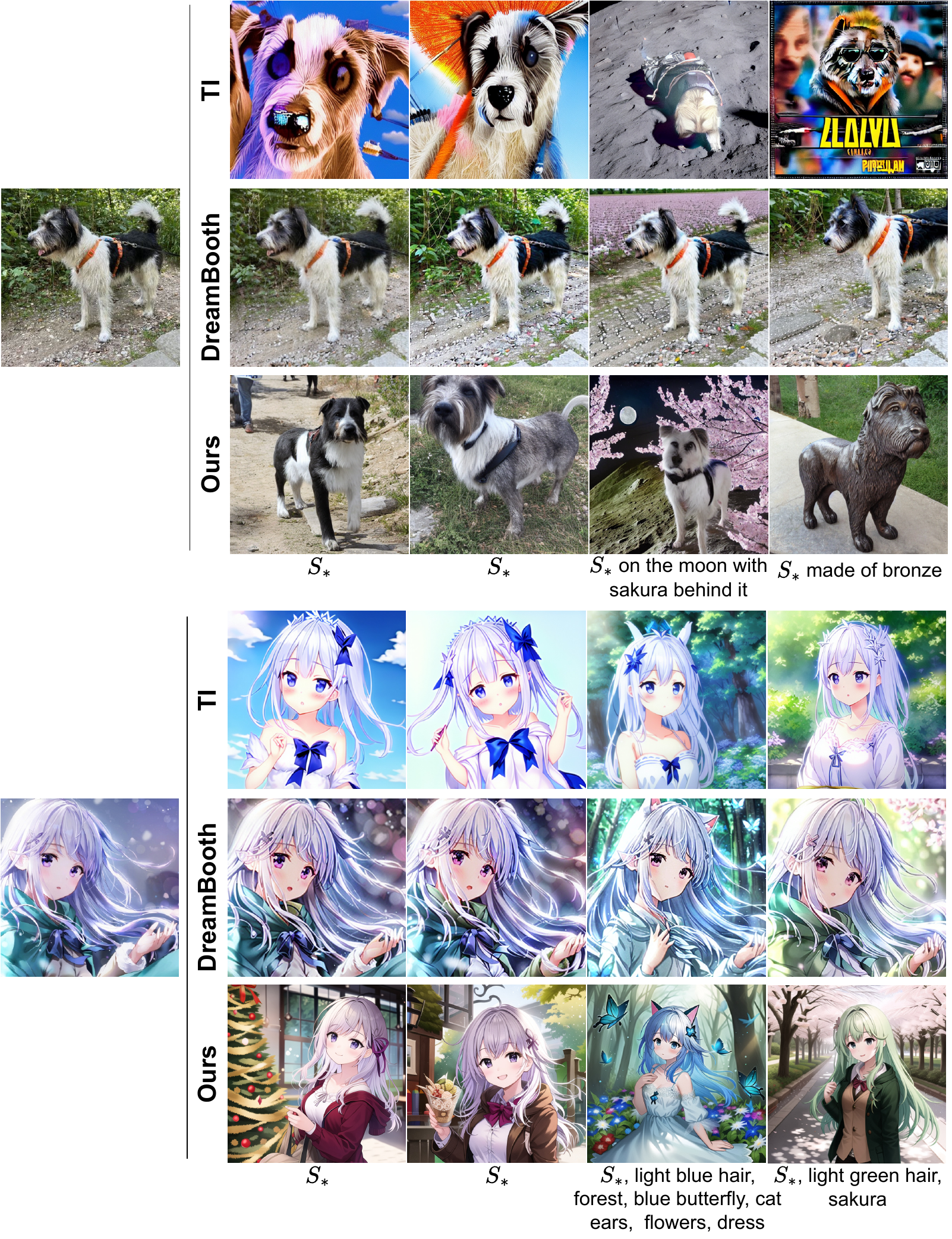}
    \vspace{-8pt}
    \caption{Comparison with existing methods for one-shot text-to-image generation.}
    \vspace{-12pt}
    \label{fig:comparison_OneShotSetting}
\end{figure}

\subsection{Deficiencies in Existing Methods}
The majority of existing text-to-image Diffusion models, including LDM, incorporate the Conditional Fine-Grained (CFG) mechanism to enhance performance.
More specifically, CFG utilizes an implicit classifier to perform guidance with $\frac{p(z_t|c)}{p(z_t)}$.
However, the influence of this mechanism on the training process has been overlooked in previous few-shot text-to-image generation methods, leading to issues such as poor controllability or low-quality generated images.

The one/few-shot generation method learns from the given reference image(s) $w \sim p(w)$, where $p(w)$ is a distribution different from the distribution of the pre-training data $p(z)$.
TI uses prompt tuning to learn a pseudo word $S^p$ and maximizes log-likelihood of $p_{\theta}(w_t|S^p)$. With the fact that $w$ and $z$ will eventually diffuse to $\mathcal{N}(0, \mathcal{I})$ at timestamp $T$, we have $w_T$=$z_T$; then $p_{\theta}(w_{T-1}|w_T,S^p)=p_{\theta}(w_{T-1}|z_T,S^p)$. Accordingly, TI generates images with $\frac{p_{\theta}(w_t|S^p)}{p_{\theta}(z_t)}$. 
However, $p(w)$ and $p(z)$ are different distributions, TI only fits $p_{\theta}(w_t|S^p)$ while ignoring the differences between $p_{\theta}(w_t)$ and $p_{\theta}(z_t)$, leading to incorrect guidance directions and generating unexpected features and artifacts. 
Moreover, in the one-shot learning tasks, when $p(w)$ forms a single point, simply maximizing the log-likelihood of $p_{\theta}(w_t|S^p)$ can lead to severe overfitting and limited diversity.
This problem is the same for fine-tuning-based methods like DB or Adapter-based methods like lora.  These approaches aim to maximize the log-likelihood of $p_{\theta}(w_t|S^p)$ by learning $\theta$, and similarly ignoring the distinctions between $p_{\theta}(w_t)$ and $p_{\theta}(z_t)$.
}

\section{Methodology}
% To overcome the limitations of existing methods aforementioned, enabling the model to synthesize highly realistic and diverse images with high controllability 
% %according to the user-given patterns with just one input image,
% through just one user-given image, our DreamArtist is proposed, shown in \cref{fig:framework}. 
% %
% Without bells and whistles, DreamArtist just introduces both positive and negative embeddings and jointly trains them with positive-negative prompt-tuning, achieving very promising generation results even in a one-shot setting.  

%%%%%
To overcome the limitations of existing methods, our proposed DreamArtist framework, depicted in \cref{fig:framework}, synthesizes highly realistic and diverse images with enhanced controllability from a single user-provided image. Without bells and whistles, DreamArtist just introduces a novel approach that incorporates both positive and negative adapters. These adapters are concurrently optimized through positive-negative tuning, which remarkably improves generation quality, even within the challenging constraints of a one-shot scenario.

%%%%%

\subsection{Positive-Negative Prompt-Tuning}
In essential, conventional prompt-tuning~\cite{PTuning2, PromptZeroshot, PromptCloze, CoPo, CoCoPo, PTVision, PromptTuning, PrefixTuning} optimistically considers only one prompt. Namely, it directly aligns it with the downstream task and learns a mapping from the prompt to the training set.
However, based on our aforementioned analysis, applying these methods directly to the diffusion model may easily lead to collapse and overfitting, particularly when the training stage involves a limited samples.
%However, this easily leads to collapse and over-fitting, when there are very few samples given in the training stage. 
Especially, for one-shot text-to-image generation, 
this prompt-tuning has a remarkable over-fitting to the reference image and generates images with limited diversity and even artifacts.
Accordingly, we propose positive-negative prompt-tuning to address these problems, 
%which learns to generate in a self-induced manner. 
%With PNPT, the model disentangles the conventional prompt tuning into two components and implicitly induces and corrects the generation, like positive-negative learning or introspection.
which disentangles the conventional prompt tuning into two components and learns to generate in a self-induced manner. 
%That is, it enables the model not only aggressively to learn the characteristics in the image, but also to rectify its mistakes following an introspection mechanism.
% Exsiting works having the similar idea in NLP, but ....

%%%%%%%%%%%
Specifically, based on the CFG mechanism, given a noise map $z_t$, it can be guided respectively by positive and negative text prompts and yields two different feature maps $z_{t-1}^p$ and $z_{t-1}^n$. It is expected that $z^p$ contains the desired characteristics of the reference image, while $z^n$ contains the characteristics we prefer to be excluded from the generated image for rectification.
%%%%%%%%%%
Our approach involves the simultaneous learnable positive and negative pseudo-words ($S^p_*$ and $S^n_*$) and maximizes the log-likelihood of $\frac{p_{\theta}(w_t|S^p_*)}{p_{\theta}(w_t|S^n_*)}$. Therefore, the loss function of DreamArtist is defined as follows:
\begin{equation}\label{equ:pnpt_loss}
\begin{aligned}
\mathcal{L}_{da} = \| \epsilon - (\epsilon_{\theta}(z_t, S_*^n) + \gamma \left( \epsilon_{\theta}(z_t, S_*^p)-\epsilon_{\theta}(z_t, S_*^n) \right)) \|^2
\end{aligned}
\end{equation}
% As stated in \cite{GLIDE}, the model can be extrapolated in the direction of $z^p$ and away from $z^n$, with the following guiding strategy:
% \vspace{-5pt}
% \begin{equation}
% \hat{z} = f_m(z^p, z^n) = z^n + \gamma (z^p-z^n),
% \label{equ:neg_prompt}
% \vspace{-5pt}
% \end{equation}
% where $\gamma$ is a weight factor and $f_m$ indicates the function of \cref{equ:neg_prompt} for simplicity.
DreamArtist learns $S^p_*$ and $S^n_*$ jointly, which ensures that the training and inference are guided in the same direction, thus eliminating artifacts and improving the quality of image generation and controllability of features. It avoids the problem of inaccurate feature generation or uncontrollable outcomes caused by incorrect guidance.

\begin{figure*}[ht!]
    \centering
    \includegraphics[width=0.98\textwidth]{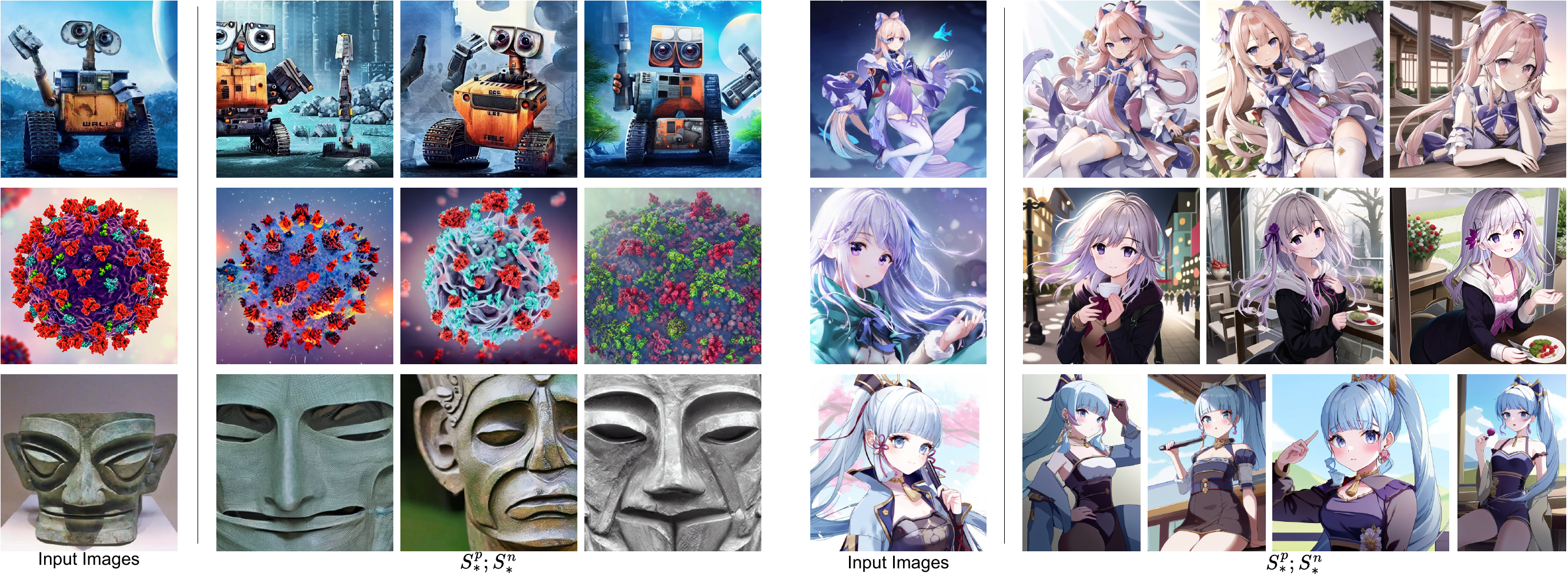}
    %\vspace{-6pt}
    \caption{One-shot text-to-image generation with only learned pseudo-words for DreamArtist. It can learn content and context from a single input image without adding additional text descriptions, generating diversity and high-quality images in both natural and anime scenes.}
    \vspace{-6pt}
    \label{fig:vis_rawSp}
\end{figure*}

In \cref{equ:pnpt_loss}, our essential optimization objective is to align the 
distribution $p(z_t|S^p_*) (\frac{p(z_t|S^p_*)}{p(z_t|S^n_*)})^{\gamma} = \frac{p(z_t|S^p_*)^{\gamma+1}}{p(z_t|S^n_*)^{\gamma}}$ with the distribution of reference image $\tilde{p}(z_t|c)$. Given that $p(z_t|S^p_*)<1$ and $p(z_t|S^n_*)<1$, we have:
\begin{equation}\label{equ:dist_align}
\left(\frac{p(z_t|S^p_*)}{p(z_t|S^n_*)}\right)^{\gamma} \geq \frac{p(z_t|S^p_*)^{\gamma+1}}{p(z_t|S^n_*)^{\gamma}} \approx \tilde{p}(z_t|c).
\end{equation}
When there is only one reference image, if only the positive branch is used, the optimization objective is to align the distribution $p(z_t|S^p_*)$ with the distribution $\tilde{p}(z_t|c)$. However, because there is only one reference image, $\tilde{p}(z_t|c)=1$, which would lead to severe overfitting and cause the model to lose controllability. In our method, the positive branch $p(z_t|S^p_*)$ does not directly align with $\tilde{p}(z_t|c)$; instead, the constraint on the positive branch is relaxed through $p(z_t|S^n_*)$. Moreover, as $\gamma$ increases, $\frac{p(z_t|S^p_*)}{p(z_t|S^n_*)}$ can become smaller, resulting in a greater degree of relaxation. Therefore, a larger $\gamma$ will lead to a lower fitting and provide higher controllability.

\subsection{Reconstruction Constraint for Detail Enhancement.}
\label{sec:rec}
% \subsection{Reconstruction Constraint}
Constraints in the feature space only, would make the generated images be smoothness and even with some deficiencies in details and colors. Thus, we add an additional pixel-level reconstruction constraint for the generated image. % to enhance the embedding's ability on describing details and colors.  
In this context, we use approximate $z_0$ for calculating the loss through the noise $\hat{\epsilon}$, which is predicted by the model based on $z_t$ through the CFG mechanism $\hat{\epsilon} = \epsilon_{\theta}(z_t, S_*^n) + \gamma \left( \epsilon_{\theta}(z_t, S_*^p)-\epsilon_{\theta}(z_t, S_*^n) \right)$.
Then the approximate $z_0$ is:
\begin{equation}
z_{t \rightarrow 0} = \frac{1}{\sqrt{\bar{\alpha}_t}}\left(z_t-\hat{\epsilon} \sqrt{1-\bar{\alpha}_t}\right),
\end{equation}
Accordingly, the reconstruction loss can be expressed as:
% \vspace{-8pt}
\begin{equation}
\mathcal{L}_{rec} = \| \mathcal{D}(z_{t \rightarrow 0}) - x \|,
\end{equation}
% \vspace{-10pt}
where $\| \cdot \|$ is the $\ell 1$ loss.

\subsection{Positive-Negative Adapter}
Due to the high dimensionality of image data, which differs from the characteristics of natural language, and the fact that humans are sensitive to the perception of details in images. Relying solely on prompt tuning for learning fine-grained features still falls short. It often struggles to faithfully reproduce the detailed features of entities present in the reference images, leading some individuals to subjectively perceive discrepancies. Additionally, limitations in controllability may also arise from insufficient comprehension of features, resulting in associations errors or difficulties in controlling details (e.g., inaccurate modification of accessories or altering hair color affecting eyes color). Moreover, if the distribution of features in the reference images significantly deviates from those in the pre-trained model, the model may struggle to learn effectively and fail to adequately characterize these features.

In response to these challenges, we propose the LoRA-based Positive-Negative Adapter method, which incorporates the Adapter. The Generator part's Adapters ($\phi^p_u$ and $\phi^n_u$) serve a dual purpose: enhancing the model's ability to preserve fine-grained details from the reference images and enabling the model to learn features more faithful that are absent in the pre-trained image data. Adapters in the Text Encoder part ($\phi^p_t$ and $\phi^n_t$) further elevate the model's controllability, allowing the features in the reference images to be precisely and finely controlled through textual descriptions.
Our DreamArtist method combines the Positive-Negative Adapter approach with prompt tuning, enabling the model to learn the generation of images with high controllability, strong diversity, intricate details, and high quality through only one reference image as input.

\section{Experiments}

\begin{figure}[!t]
    \centering
    \includegraphics[width=0.48\textwidth]{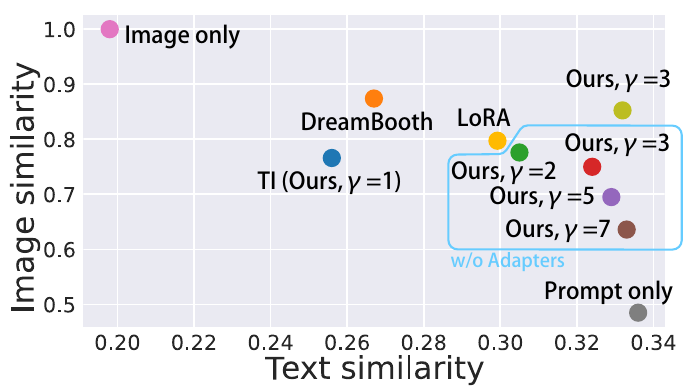}
    %\vspace{-10pt}
    \caption{Comparison of our DreamAritist with different values of $\gamma$ and existing methods, following the same evaluation method to TI.}
    %\vspace{-12pt}
    \label{fig:exp_TI_metric}
\end{figure}

\begin{table}[]
\renewcommand\arraystretch{1}
%\fontsize{8.8pt}{\baselineskip}\selectfont
\setlength\tabcolsep{10pt}

\caption{Quantitative comparison of our DreamArtist with existing methods for one-shot text-to-image generation\textsuperscript{\ref{foot:blue}}. %The TI results on EG and SMD2 are due to artifacts and thus marked in \textcolor{gray}{gray}.
%\pengxuc{remember to remove two metrics}
}

\resizebox{\columnwidth}{!}{
\begin{tabular}{|c|cccc|}
% \toprule
\hline
\multicolumn{1}{|c|}{Method}     & LPIPS↓          & Style loss↓        & CDS↑    & CFV↑      \\ 
\hline 
\rowcolor{gray!20}
\multicolumn{5}{|c|}{Natural Image Generation}                                                                                                                    \\ \hline
\multicolumn{1}{|c|}{TI}         & 0.71           & 6.12           & 0.73          & \textbf{1.79} \\
\multicolumn{1}{|c|}{DreamBooth} & \textbf{0.33} & \textcolor{blue}{1.31}   & 0.63          & 0.69          \\
LoRA & 0.47 & 1.60            & 0.48          & 0.89          \\
\multicolumn{1}{|c|}{Ours(DreamArtist w/o Adapter)}       & 0.62            & 2.46          & \textcolor{blue}{0.74} & \textcolor{blue}{1.53}  \\
\multicolumn{1}{|c|}{Ours(DreamArtist)}       & \textcolor{blue}{0.46}            & \textbf{1.29}          & \textbf{0.75} & 1.46  \\
%\hline 
\hline \rowcolor{gray!20}
\multicolumn{5}{|c|}{Anime Image Generation} \\ \hline
TI         & 0.63          & 7.47            & 0.41          & 0.87          \\
DreamBooth & \textcolor{blue}{0.49} & 1.16            & 0.33          & 0.72          \\
LoRA & \textbf{0.48} & 0.71            & 0.22          & 0.66          \\
Ours(DreamArtist w/o Adapter)       & 0.60          & \textcolor{blue}{0.69}    & \textbf{0.60} & \textbf{1.28} \\
Ours(DreamArtist)       & \textbf{0.48}          & \textbf{0.38}    & \textcolor{blue}{0.59} & \textcolor{blue}{1.03} \\
% \bottomrule
\hline
\end{tabular}
}

\label{tab:exp_base}
\end{table}

\subsection{Experimental Settings}
\noindent
\textbf{Dataset.} Following the existing experimental settings~\cite{TI, DreamBooth}, the LAION-2B dataset~\cite{LAION} is used for natural image generation. Additionally, we add an anime dataset, Danbooru~\cite{danbooru2021} for the popular interest in many applications, \eg, games and animes. 

\noindent
\textbf{Implementation details.}
In DreamArtist, the learning rate is 0.0025 and $\gamma$ is 3 (5 for style cloning). It is trained on one RTX2080ti using a batch size of 1 with about 2k-8k iterations. 
% (the optimal number of iterations is different for different data). 
% In comparison with similar methods, 
In TI, the length of an embedding for a pseudo-word is set to equal that of 6 words, while DreamArtist uses 3 words for both positive and negative embeddings.
\revise{Positive prompt is initialized using similar words similar to TI (initialized with some similar words), while negative prompt is initialized using EOS token (referring to empty text $\emptyset$) with random noise. In the CFG configuration, the negative prompt is by default set to empty text. The negative prompt is derived from $p(z_t)$, so we use empty text with a small noise to initialize the negative embedding.}

\begin{table}[]
\centering
\footnotesize
\renewcommand\arraystretch{1.1}
%\fontsize{8.8pt}{\baselineskip}\selectfont
\setlength\tabcolsep{1pt}
\caption{Quantitative analysis of our DreamArtist compared with existing methods on feature controllability. The feature controllability of DreamArtist substantially exceeds existing methods\tablefootnote{Values in \textbf{bold} are the best results and those in \textcolor{blue}{blue} are the second best. \label{foot:blue}}.}

\resizebox{1.02\columnwidth}{!}{
\hspace{-2ex}
\begin{tabular}{|c|cccc|cccc|}
\hline 
  \multirow{2}{*}{Method}   & \multicolumn{4}{c|}{Natural Image Generation}                                                              & \multicolumn{4}{c|}{Anime Image Generation}                                         \\ 
     & CAS↑           & CFV↑          & Style loss↓    & CDS↑        & CAS↑           & CFV↑          & Style loss↓    & CDS↑        \\ \hline
TI         & 0.37          & \textcolor{blue}{1.46}          & 4.31         & \multicolumn{1}{c|}{0.40}          & 0.23          & 0.98          & 5.52          & 0.46          \\
DreamBooth & 0.24          & 1.19          & \textbf{0.40} & \textbf{0.69} & 0.28          & 0.81          & 1.28 & 0.31          \\
LoRA & 0.30          & 1.10          & 1.02 & 0.53 & 0.37          & 0.80          & \textcolor{blue}{1.03} & 0.13          \\
Ours(DreamArtist w/o Adapter)       & \textcolor{blue}{0.83} & \textbf{1.55} & 2.01          & \multicolumn{1}{c|}{0.57}          & \textcolor{blue}{0.63} & \textbf{1.15} & 2.71          & \textbf{0.58} \\ 
Ours(DreamArtist)       & \textbf{0.89} & 1.43 & \textcolor{blue}{0.98}          & \multicolumn{1}{c|}{\textcolor{blue}{0.58}}          & \textbf{0.65} & \textcolor{blue}{1.07} & \textbf{0.91}          & \textcolor{blue}{0.57} \\ \hline
\end{tabular}
}
%0.373 	0.797 	1.025 	0.132 

%\vspace{-4ex}
\label{tab:exp_ctrl}
\end{table}

\begin{table}[]
\centering
\footnotesize
\renewcommand\arraystretch{1.0}
%\fontsize{8.8pt}{\baselineskip}\selectfont
\setlength\tabcolsep{5pt}
\caption{Evaluation on $\gamma$ and $\mathcal{L}_{rec}$\textsuperscript{\ref{foot:blue}}.}

\resizebox{1.0\columnwidth}{!}{
\begin{tabular}{|c|cccc||c|}
\hline
 Method  & LPIPS↓ & Style loss↓ & CDS↑  & CFV↑ & CAS↑  \\ \hline
$\gamma$=3; w/o $\mathcal{L}_{rec}$ & 0.629 & 1.41       & 0.68  & 1.33  & 0.76 \\ \hline
$\gamma$=2               & \textbf{0.601} & \textbf{1.13}       & 0.59 & 1.02  & 0.67 \\
$\gamma$=3               & \textcolor{blue}{0.613} & \textcolor{blue}{1.17}       & \textbf{0.72}  & \textcolor{blue}{1.41}  & 0.79 \\
$\gamma$=5               & 0.640 & 1.85       & 0.48  & \textbf{1.66}  & \textcolor{blue}{0.87} \\
$\gamma$=7               & 0.653 & 1.44       & \textcolor{blue}{0.67}  & 1.03  & \textbf{0.92} \\
\hline
\end{tabular}
}

\label{tab:exp_ab}
\end{table}

\noindent
\textbf{Metrics.}
\textbf{1)} For quantitative evaluation, we follow the same evaluation metrics to TI, namely calculating \emph{image similarity} to the \textbf{given} reference images and \emph{text similarity} to the \textbf{given} texts. But, this presents a \textbf{measure bias} to the overfitting model to the \textbf{given} reference images and texts, especially for one/few-shot generation (\cref{sec:one_shot_exp}). 
%Besides, these are applicable for generation given additional text descriptions, but not for evaluating the single learned pseudo-word. 
\textbf{2)} Accordingly, we also adopt different metrics from three aspects\footnote{
More details on CDS, CFV and CAS are described in Supplementary.}:  
image/style similarity (LPIPS~\cite{LPIPS}/style loss~\cite{styloss}), 
image diversity (CLIP detail score, CDS; CLIP image feature variance, CFV), and generation controllability (CLIP average score, CAS). \emph{Notably, there is a tradeoff between the image diversity and similarity to given reference images.} High \textbf{similarity} indicates a potential overfitting bias and a limited ability of generating \textbf{diverse} images, and vice versa. 

\textbf{CDS} employs the CLIP model to evaluate the richness of the details of the generated image content. 
%is based on the CLIP model and evaluates how detailed the images are. 
More precisely, it represents 
the probability of the image being categorized as "detailed" within the set of [little detail'', detailed''] using CLIP.
%the probability that CLIP assigns to the image being categorized as "detailed" within the set of [little detail'', detailed''].

%\noindent
\textbf{CFV}
is used to evaluate the diversity of generated images.  %This metric operates by 
It calculates the standard deviation of the feature maps of the generated images. Those feature maps are encoded through the image encoder of CLIP.

%\noindent
\textbf{CAS}
is used to evaluate the generation controllability, to check whether additional text descriptions are accurately rendered in the generated image. 
Firstly, we extract all noun phrases from the text descriptions. Secondly, each of these noun phrases is fed into the CLIP model together with a randomly selected set of noun phrases from the database. Notably, those selected noun phrases do not exist in the text descriptions. Then, the average of the probabilities that CLIP classify 
the generated image to the noun phrases in the descriptions is CAS.

\begin{figure*}[ht!]
    \centering
    
    \includegraphics[width=0.98\textwidth]{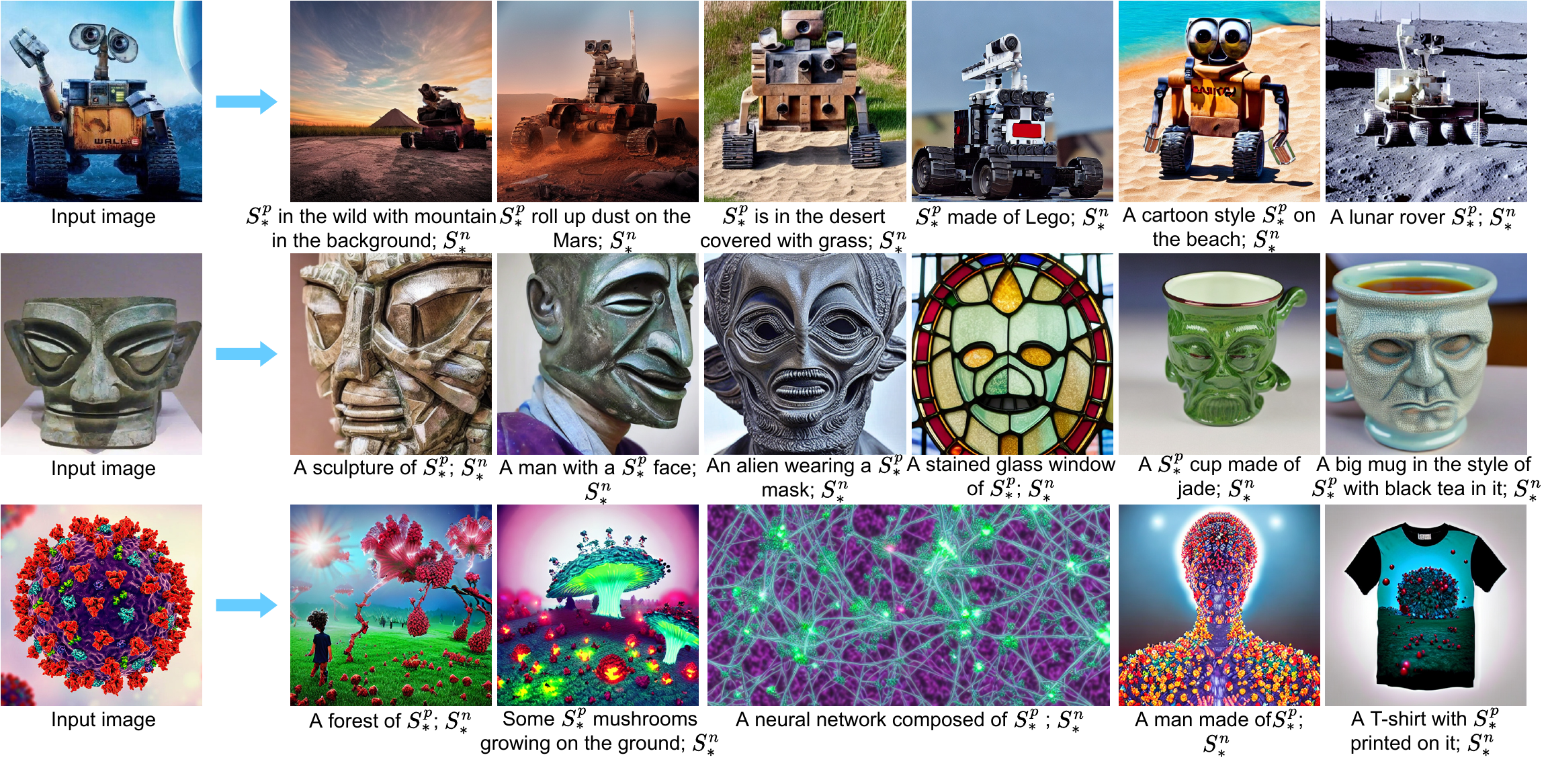} \\
    \vspace{-8pt}
    \hspace{-8pt}
    \includegraphics[width=0.97\textwidth]{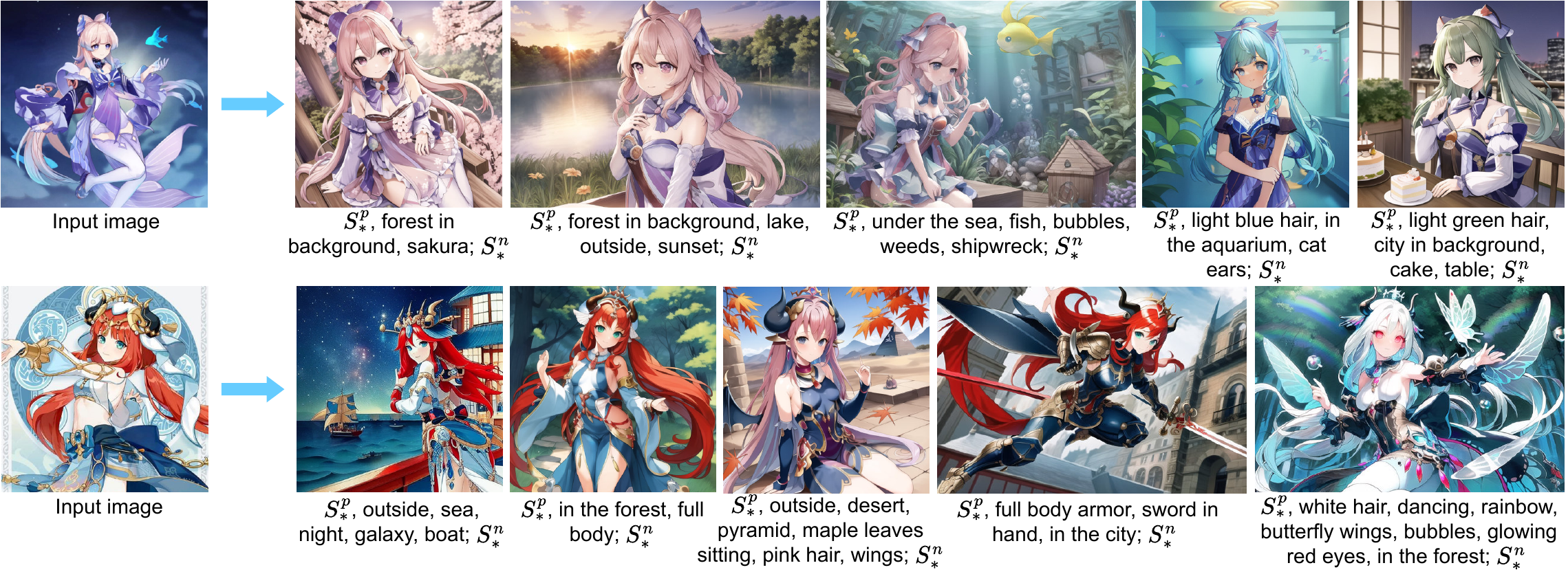}
    %\vspace{-9pt}
    \caption{
    One-shot text-to-image generation with the guidance of additional complex texts for DreamArtist.
    %The generated images of DreamArtist with the guidance of additional complex texts.
    DreamArtist exhibits a superior capability of controllable generation: even with few words in the text guidance, diverse and faithful images are generated; with more words, vivid images with rich details are generated. More importantly, %DreamArtist can be good at realizing almost all the given words.
    DreamArtist can successfully render almost all the given words.
    }
    \vspace{-8pt}
    \label{fig:vis_ComplexTexts}
\end{figure*}

%\subsection{Learning new concepts to represent a set of features}
\subsection{One-Shot Text-guided Image Synthesis}
\label{sec:one_shot_exp}
We compare our DreamArtist with two existing works, including TI~\cite{TI}, DreamBooth~\cite{DreamBooth} and LoRA~\cite{LoRA} for one-shot text-to-image generation\footnote{Their generated images given 3-5 images are shown in \cref{fig:comparison_TI_DB} and more results are provided in Supplementary.}. All methods are trained with only one image given as reference for a fair comparison in all the experiments. 
Next, we will elaborate the comparison results from image similarity and diversity, generation controllability, and style cloning. 

\begin{figure*}[t!]
    \centering
    
    \includegraphics[width=0.98\textwidth]{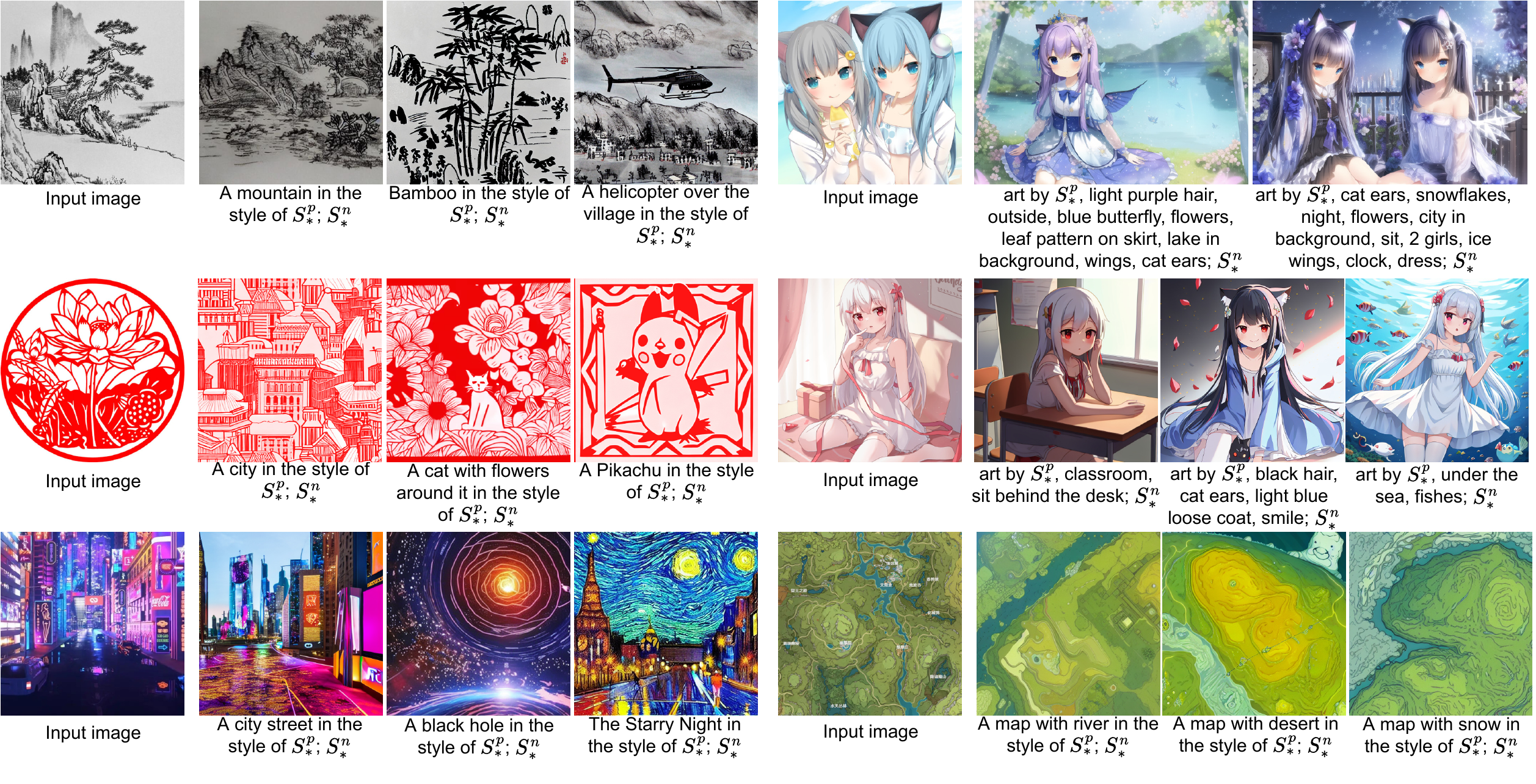}

    \vspace{-6pt}
    \caption{Style cloning via DreamArtist, for example, styles of wash painting,  paper-cut art, Cyberpunk, comic of caricaturists, and road map in a game (from left to right). 
    % not only learns characteristics, but also some realistic styles. In anime scenes it is also possible to clone the drawing style of an artist. 
    % Recent  methods can only learn some abstract art styles.
    }
    \vspace{-6pt}
    \label{fig:vis_style}
\end{figure*}

\begin{figure*}[ht!]
    \centering
    \includegraphics[width=0.98\textwidth]{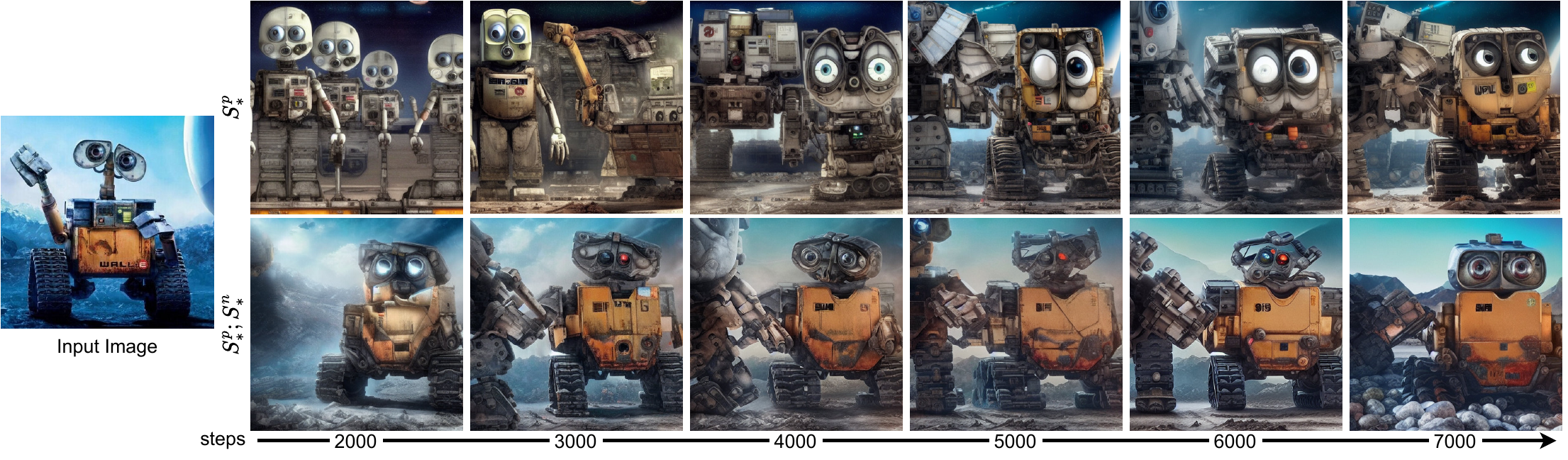}
    %\vspace{-10pt}
    \caption{
    Generation results via only $S^p_*$ or ($S^p_*$;$S^n_*$) at different steps. UP: using $S^p_*$ alone. Down: using an $S^p_*$ after 7000 steps together with $S^n_*$ from different steps. It is observed that $S^n_*$ rectifies the generation based on $S^p_*$ to improve the image quality.
    % Up: The generated results of $S^p_*$ alone at each iteration step. Down: The generated results using 7000 steps of $S^p_*$ with $S^n_*$ at different iteration step.
    }
    \vspace{-10pt}
    \label{fig:vis_evolution}
\end{figure*}

\noindent
\textbf{Image Similarity and Diversity.} 
Qualitatively, \cref{fig:comparison_OneShotSetting} shows that images generated by TI have  limited diversity and details. DreamBooth generates images with fairly quality, but the diversity is also limited for both natural and anime image generation. It generates images overly similar to the reference image, which evidences an over-fitting issue. From \cref{fig:comparison_OneShotSetting} and \ref{fig:vis_rawSp}, our DreamArtist can alleviate these problems and not only generates highly realistic images with more promising and reasonable details, but also keeps the generated images highly diverse, \eg, dogs in different contexts (\cref{fig:comparison_OneShotSetting}) and WALL-Es with difference appearances (\cref{fig:vis_rawSp}).

Quantitatively, we first follow the same evaluation method of TI and provide the comparison in \cref{fig:exp_TI_metric}. Our DreamArtist (Ours, $\gamma=3$) has a superior balance of image generation between image similarity and text similarity, in comparison with TI and DreamBooth. This is also demonstrated in \cref{tab:exp_base} and \cref{tab:exp_ctrl}. Though DreamBooth and LoRA performs better on LPIPS and style loss due to over-fitting, it has a poor diversity (CFV). TI has a higher CFV for the diversity, but suffers from severe artifacts that make an illusion of high diversity metrics, as demonstrated in \cref{fig:comparison_OneShotSetting}.

\noindent
% \textbf{Image Quality and Diversity.}
\textbf{Generation Controllability.}
As shown in \cref{fig:comparison_OneShotSetting}, TI has a limited generation controllability that it fails to render some of other words, \eg green hair. Text similarity in \cref{fig:exp_TI_metric} and CAS in \cref{tab:exp_ctrl} also demonstrate that. Although DreamBooth can render some additional texts, the generated images are too homogeneous in structure and extremely poor in diversity. This also indicates the limited controllability of DreamBooth. 
In contrast, DreamArtist can keep high controllability and diversity while maintaining sufficient similarity to the reference image. Besides, from CAS in \cref{tab:exp_ctrl}, DreamArtist also performs best with a significant improvement (0.89 [DreamArtist] vs. 0.24/0.37 [DreamBooth/TI]).

\begin{figure*}[ht!]
    \centering
    \includegraphics[width=0.98\textwidth]{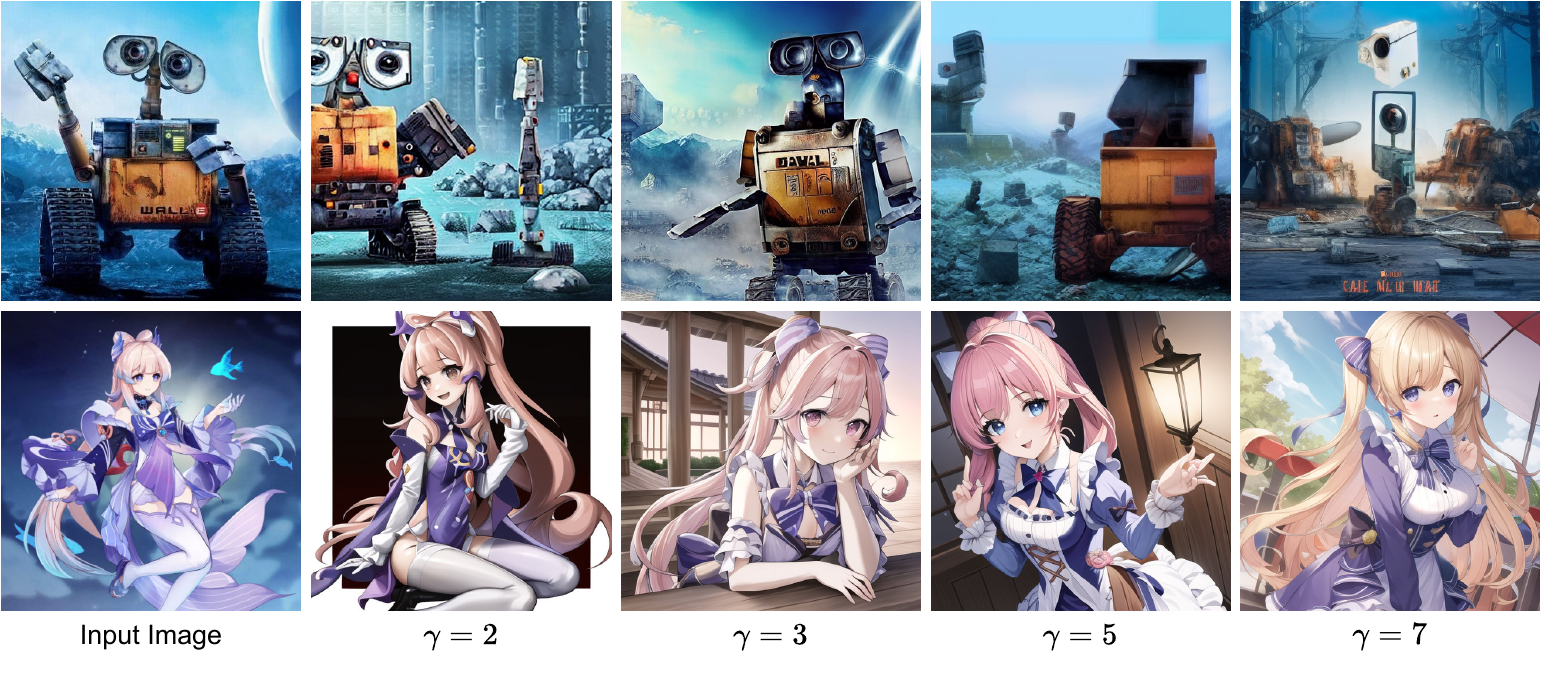}
    %\vspace{-10pt}
    \caption{The visualization results of training with different values of $\gamma$. }
    \vspace{-10pt}
    \label{fig:exp_gamma}
\end{figure*}

\begin{figure*}[ht!]
    \centering
    \includegraphics[width=0.98\textwidth]{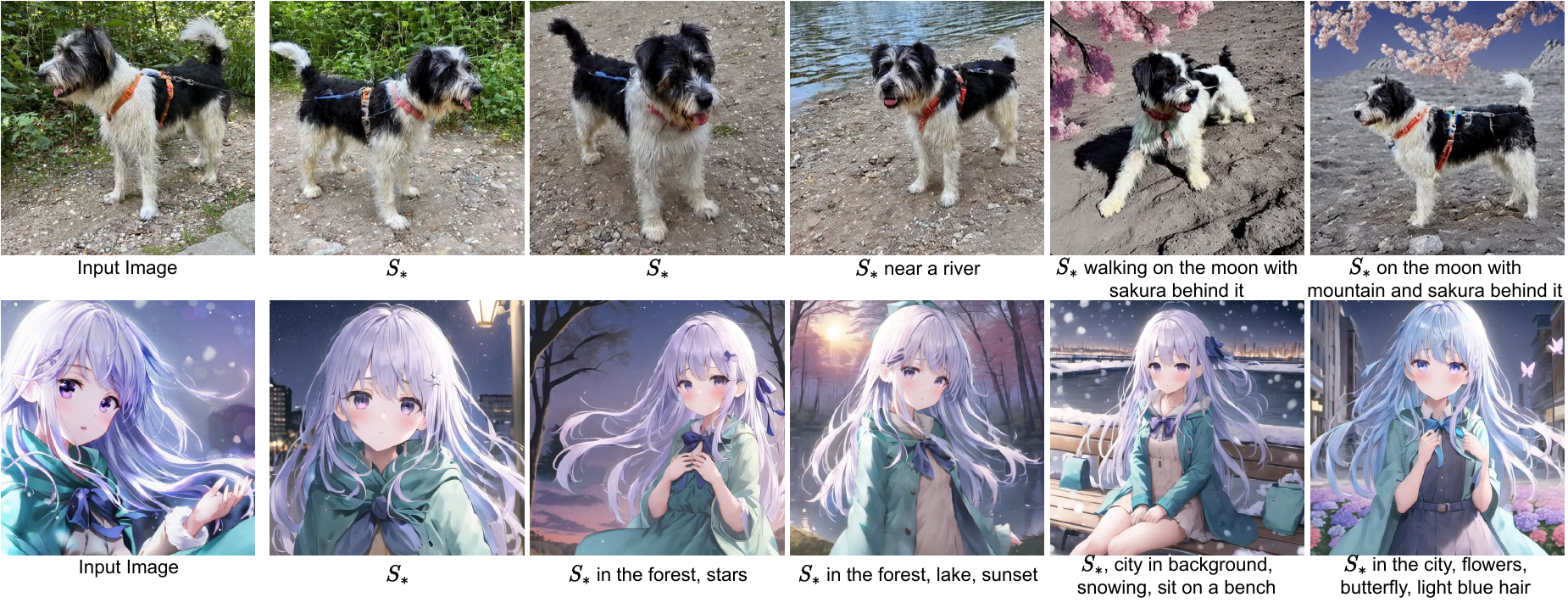}
    %\vspace{-10pt}
    \caption{Visualization results DreamArtist with LoRA. Compared to DreamArtist, which relies only on prompt tuning only, the detail and faithful of the generated image to the reference image is significantly improved by adding an adapter.}
    \vspace{-10pt}
    \label{fig:exp_dapp}
\end{figure*}

\vspace{2pt}
\noindent
\textbf{Style Cloning.}  
In \cref{fig:vis_style}, it is observed that DreamArtist can learn different image styles, \eg, Chinese brush painting, paper-cut painting and Cyberpunk.
In anime cases, different artists have different painting styles of different brushwork, composition, light processing, color processing, scenery, and many other details. The different painting styles are not as diverse as the different styles in natural scenes, but they will give the audience a different impression. Our method can learn a painting style fairly well.
%, the drawn images will show a highly consistent painting style of the training image. 
It is even possible to create images that are highly similar to other works by the same artist based on the text description, which is very promising for the creation of anime work. Besides, DreamArtist also manages to generate different game maps that seem reasonable in \cref{fig:vis_style}.

\section{Method Evaluation and Analysis}

\subsection{Evaluation on $\gamma$} 
$\gamma$ controls a generation trade-off between the image similarity and controllability in comparison with the reference image.  
In \cref{tab:exp_ab}, the smaller the $\gamma$, the better DreamArtist performs for its LPIPS and style loss. This indicates the generated images are more faithful to the input image, but also underlies the inferior generation controllability and image diversity. This is demonstrated by its lower performance of CFV and CAS. On the contrary, the smaller the $\gamma$, the better DreamArtist performs for its CFV and CAS. This verifies the improvement of its generation controllability and image diversity. 
In our paper, we use $\gamma=3$ to report empirical results ($\gamma=5$ for learning image styles). 

\begin{figure*}[ht!]
    \centering
    \includegraphics[width=0.9\textwidth]{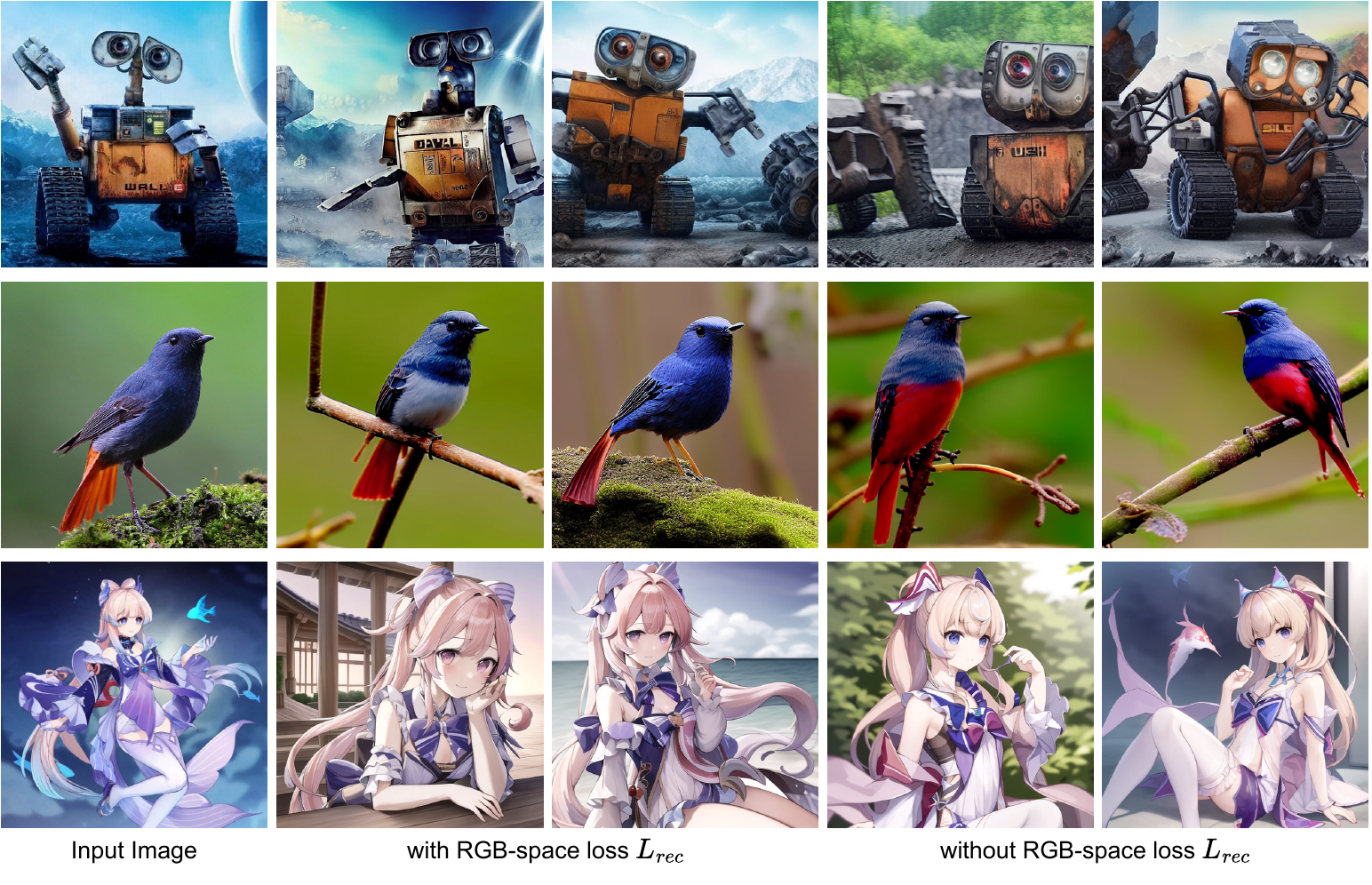}
    \vspace{-5pt}
    \caption{Visualization results of ablation study on $L_{rec}$.}
    \vspace{-10pt}
    \label{fig:exp_rec}
\end{figure*}

\begin{figure*}[ht!]
    \centering
    \includegraphics[width=0.9\textwidth]{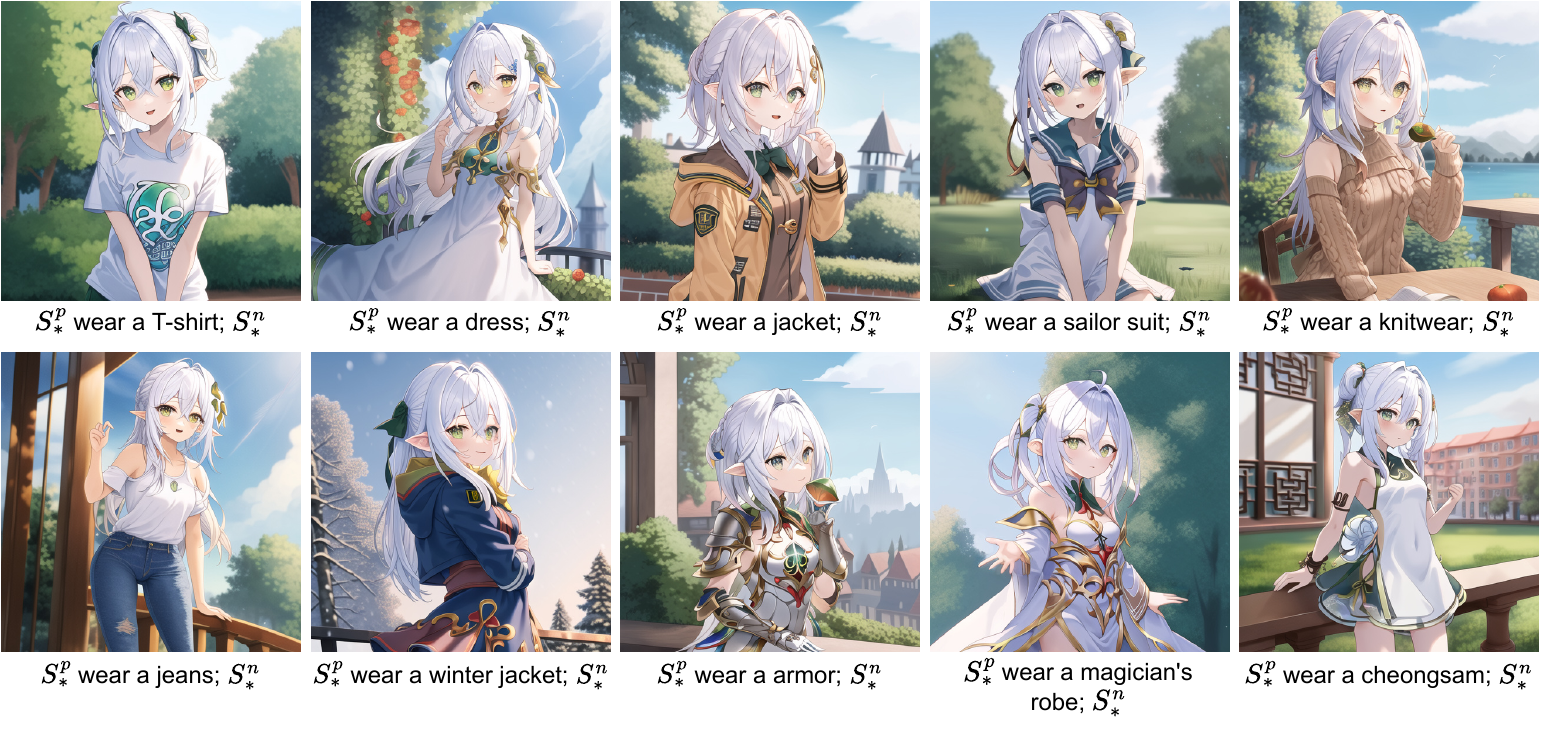}
    \vspace{-5pt}
    \caption{Generation controllability evaluation. Given different texts on clothing styles, our DreamArtist can produce an image of a girl with different styles of clothes.}
    \vspace{-10pt}
    \label{fig:exp_clothes}
\end{figure*}

% \textbf{Ablative Study. on Reconstruction Constraint.}
\subsection{Ablative Study} 
DreamArtist is very simple but effective. Its main component is positive-negative adapter, without which TI can be regarded as its base model. Thus its ablation study is identical to the comparison of TI and DreamArtist, discussed in \cref{sec:one_shot_exp}. 
Moreover, we also provide the ablation study on the reconstruction constraint (\cref{sec:rec}) in \cref{tab:exp_ab} and \cref{fig:exp_rec}. We observe that, without it, the model has difficulty learning low-level features such as color and details (\eg, the background color of WALL-E and the hair color of some anime characters. The bird's tail color is wrongly diffused to the body, which is different from the reference image.) when using only the feature space loss. 

%DA++
\noindent
\textbf{Ablation on Positive-Negative Adapters}
We also compared the performance between DreamArtist without Adapters, which utilizes only prompt tuning, and DreamArtist, which incorporates with adapters. As depicted in \cref{fig:exp_dapp}, DreamArtist exhibits a significant improvement in its ability to learn fine details from the reference image and preserve the primary features of the reference image well. 
Furthermore, both image quality and feature controllability experience a noticeable enhancement. For instance, modifications made to features such as hair color do not affect other elements like clothing.
Despite achieving better preservation of reference image features, our DreamArtist does not demonstrate signs of overfitting, maintaining commendable diversity and controllability in the generated images. The visual attributes and characteristics of the entities depicted in the images remain controllable through textual descriptions.

\begin{figure*}[t!]
    \centering
    \includegraphics[width=0.98\textwidth]{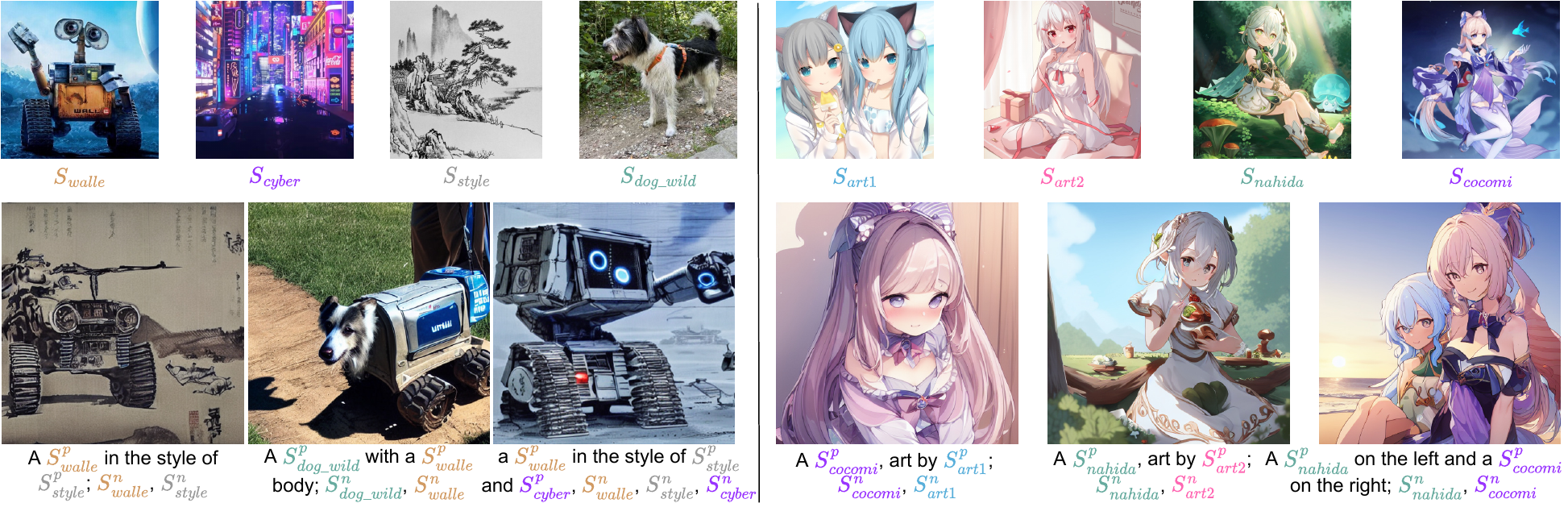}
    \vspace{-5pt}
    \caption{Results of concept compositions via DreamArtist. It presents a promising generation potential via the text guidance from the combination of the learned pseudo-words.
    % It generates images on mixing content with style, content and content, or two styles. The pseudo-words it learns can be combined at will like the original words of the generative model.
    }
    \vspace{-5pt}
    \label{fig:vis_composition}
\end{figure*}

\begin{figure}[t!]
    \centering
    \includegraphics[width=0.48\textwidth]{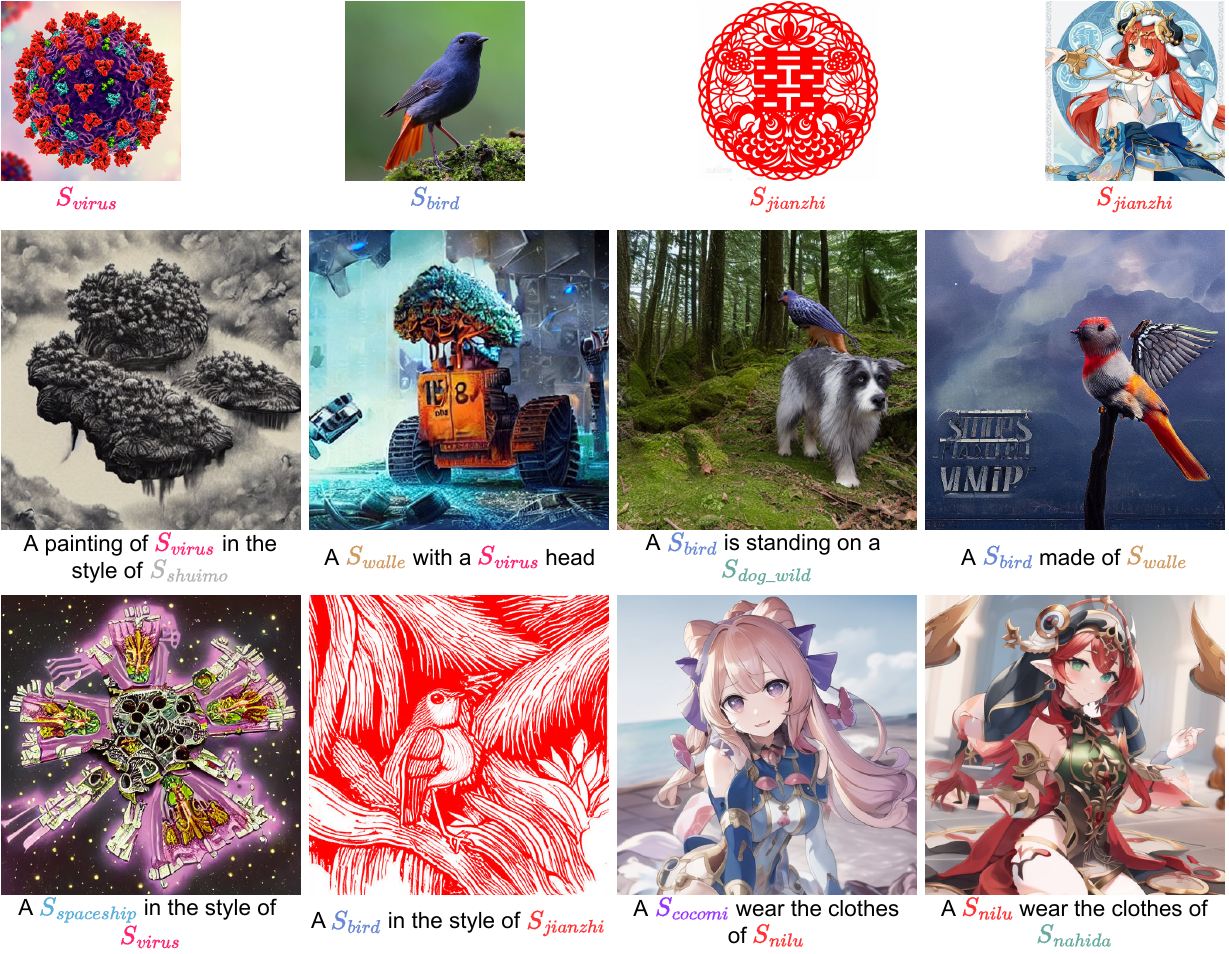}
    \vspace{-10pt}
    \caption{More results of concept compositions via DreamArtist.
    }
    \vspace{-5pt}
    \label{fig:vis_composition2}
\end{figure}

\begin{figure}[ht!]
    \centering
    \includegraphics[width=0.4\textwidth]{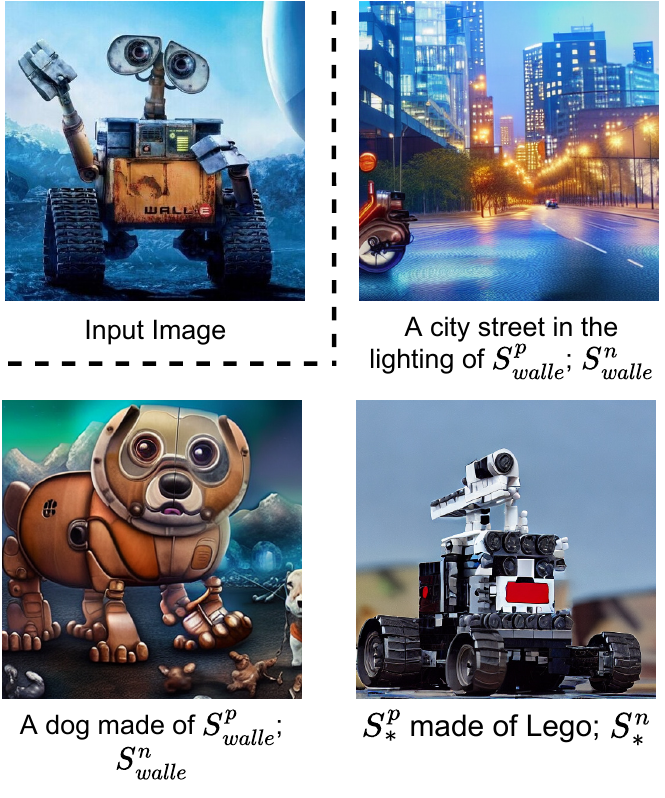}
    \vspace{-5pt}
    \caption{Visualization results of concept decoupling. DreamArtist can apply abstract semantic concepts from reference images to a given context.}
    \vspace{-10pt}
    \label{fig:exp_decopule}
\end{figure}

% \textbf{Model Learning Analysis.}
% \textbf{Positive-negative Prompt-tuning Analysis.}
\subsection{Analysis on Negative Branch}
To further explain the mechanisms of $S^p_*$ and $S^n_*$, we separately visualize the images generated by only $S^p_*$ at different steps, and the images generated by a fixed $S^p_*$ (trained for 7000 steps) together with $S^n_*$ at different steps. In \cref{fig:vis_evolution}, it is observed that $S^p_*$ steadily learns the salient features of the reference image. %, but not excessively similar and thus maintaining diversity. 
But the quality of the images generated by $S^p_*$ is not so satisfactory, \eg, lacking in style and details. 
% lack in style and details. 
When resorting to the help of $S^n_*$, the model rectifies deficiencies of $S^p_*$ and progressively improves the image quantity. This indicates the effectiveness of PNPT.

\noindent
\textbf{Analysis on Controllability and Compatibility.}
% 基于attention的分析
As discussed in \cref{sec:one_shot_exp} and demonstrated in \cref{fig:comparson_attention}, it is challenging that the learned pseudo-words of TI  harmoniously combine with other words. Namely, TI would ignore some of words in the generated images. DreamBooth has slightly better compatibility but suffers from overfitting of image content and thus has inferior image diversity.  
Our method, as an alternative, can effectively address these issues.
DreamArtist is highly compatible with complex descriptions and can generate diverse and harmonious images using learned features. The learned embedding, for instance, a mask in the second row, can produce highly realistic and diverse images based on various complex descriptions. 
Our method can also handle conflicts between the additional description and the learned features, as demonstrated in the fourth row of \cref{fig:vis_ComplexTexts}, where DreamArtist generates characters with light green hair and a city background despite the training image having pink hair and a pure background.

\subsection{Analysis on Concept Decoupling}
DreamArtist can learn not only the entities in the reference image, but also many different characteristics (concepts) such as light, material, and style (in \cref{fig:exp_decopule}). Namely, the learned pseudo words with adapters in different text contexts can be rendered with different desired characteristics of the reference image. This indicates the implicit decoupling of characteristics or concepts for promising generation controllability.

% \begin{figure*}[t!]
%     \centering
    
%     \includegraphics[width=0.98\textwidth]{imgs/exp4_edit_.pdf}

%     \vspace{-10pt}
%     \caption{Text-guided image editing via DreamArtist.
%     % The embedding learned by the DreamArtist method can also be used for text-guided image editing as well as in stable diffusion, which can generate images highly harmonious with the context.
%     }
%     \vspace{-8pt}
%     \label{fig:vis_edit}
% \end{figure*}

\subsection{Human Evaluation}
To demonstrate that our method can synthesize high-quality realistic images,
%we conducted a Turing test with 700 subjects for the TI and DPI methods, respectively. 
we have conducted a user study on TI and our methods following the rules of the Turing test from 700 participants, respectively. For testing one method, 
its 12 generated images and other 8 real images are provided to participants, to select which images are real, not generated. 
TI achieves a failure rate of 26.6\%.
%, which is unable to pass the Turing test. 
Instead, our method is 34.5\%, which significantly exceeds the Turing test requirement of 30\%. %And the DPI method even makes the subjects get an error rate of 39.8\% on virus and dog images.
This shows that the images generated by our method are fairly realistic and difficult to be distinguished from the real images.

Besides, to evaluate which creation has higher quality, 52.31\% and 83.13\% of participants from various walks of life (even including professional anime artists) prefer the synthesized images of DreamArtist for natural image cases or anime cases, respectively.

{\Rcolor
\subsection{Analysis on Embedding Length}
\begin{figure*}[ht!]
    \centering
    \includegraphics[width=0.9\textwidth]{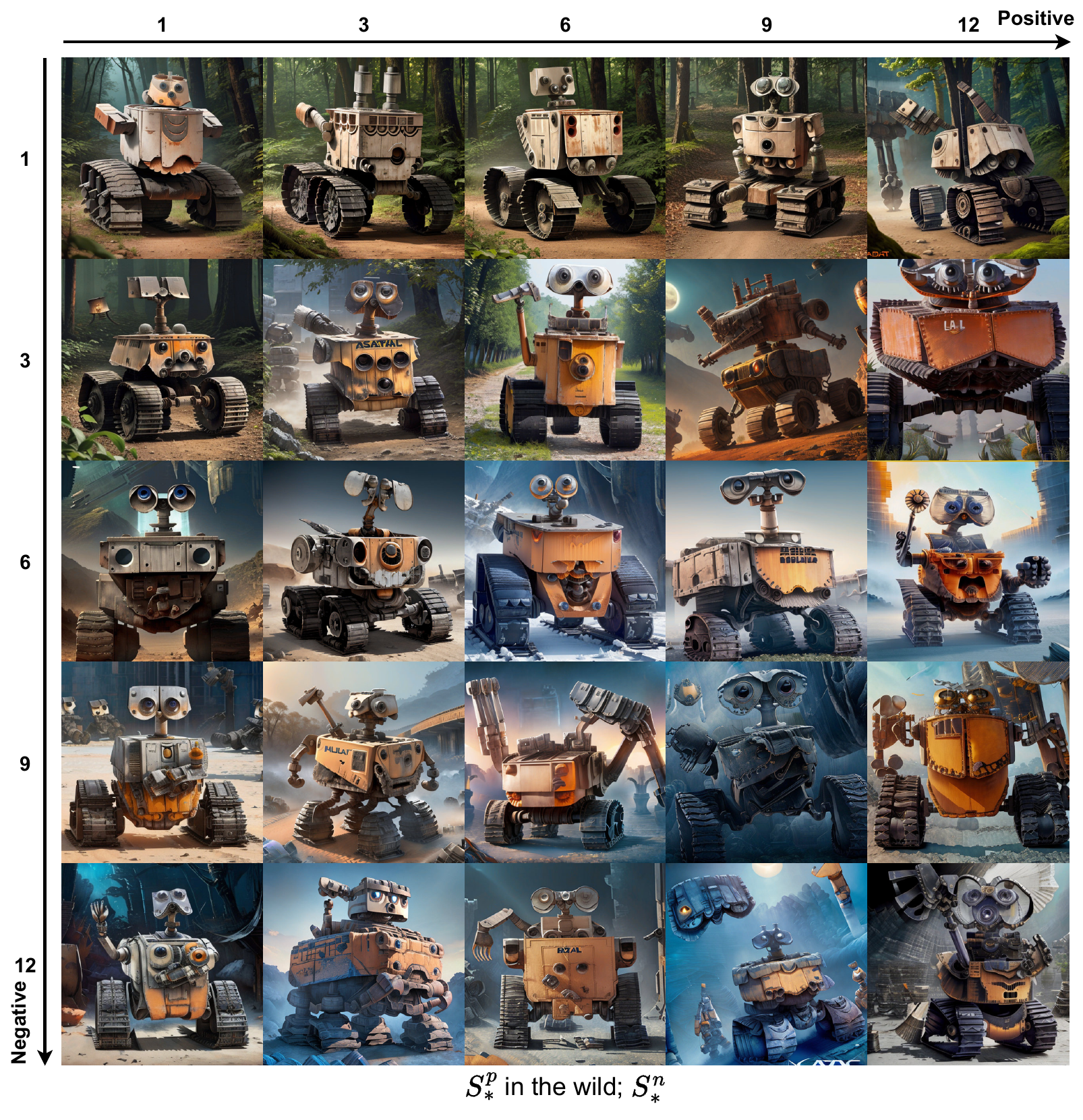}
    \vspace{-5pt}
    \caption{Visualization results of DreamArtist with different positive and negative embedding length.}
    \vspace{-10pt}
    \label{fig:exp_embnum}
\end{figure*}

As illustrated in \cref{fig:exp_embnum}, on one hand, $S^p_*$ with a length of 1 captures only coarse features, which limits the model's ability to learn finer details such as color and texture. Increasing the $S^p_*$ length to 3 achieves a more effective balance between the learned visual features and the textual description, leading to improved results. However, further extending the $S^p_*$ length beyond this point induces overfitting, causing the model to over-emphasize features from the reference image. This, in turn, compromises the alignment between the generated images and the textual descriptions, while also introducing visual artifacts.

On the other hand, increasing the $S^n_*$ length generally enhances image quality, particularly in terms of texture refinement, lighting effects, and detail preservation. Nevertheless, excessively long $S^n_*$ can also degrade image quality. Based on these empirical observations, we identify an optimal configuration of 3 for the $S^p_*$ length and either 3 or 6 for the $S^n_*$ length. To maintain consistency and simplicity, we adopt a length of 3 for both positive and negative embeddings in our work.

}

\revise{\subsection{Extended Task: Concept Compositions}}
%TI method has mentioned in their paper that TI method is struggling to combine multiple pseudo-words, while our method solves this problem well.
Our method can easily combine multiple learned pseudo-words, not only limited to combining objects and styles, but also using both objects or styles, for generating reasonable images. When combining these pseudo-words, it is necessary to add their learned embeddings of the positive and negative prompts. 
As illustrated in \cref{fig:vis_composition}, combining multiple pseudo-words (highlighted in different colors) trained with our method show excellent results in both natural and anime scenes. 
%being able to include both parts of the features at the same time, even combining two radically different objects or styles. 
Each component of the pseudo-words can be rendered in the generated image. 
% , even combining two \pengxu{radically different} objects or styles. 
For example, we can have a robot painted in the style of an ancient painting, or make a dog have a robot body. 
These are difficult to realize for existing methods. For example, in the work of TI, it mentions that TI is struggling to combine multiple pseudo-words~\cite{TI}.

%\subsection{Prompt-guided image editing}
% \subsection{Extended Task 2: Prompt-Guided Image Editing}
% As seen from \cref{fig:vis_edit}, the pseudo-words learned by our method work well for text-guided image editing, which follows the similar way of LDM~\cite{LDM}.
% % The modified areas not only show the learned \pengxu{form and content}, but also integrate well into the context, which looks harmonious. 
% The generated parts integrate well into the context, which looks harmonious. 
% The performance of image editing with the learned features of our method is as effective as that of the original features from LDM.

\section{Conclusions}
We introduce a one-shot text-to-image generation task, using only one reference image to teach a text-to-image model to learn new characteristics.
%to learn a pattern with a single image. 
Existing methods not only require 3-5 reference images, but also suffer from over-fitting that is adverse to the image diversity and generation controllability.
To mitigate this issue, we propose a simple but effective method, named DreamArtist,  without bells and whistles.
It employs a learning strategy of positive-negative prompt-tuning, enabling the model to learn and rectify the generation results. 
%to learn the decoupled features. 
%The DPI method can decouple individual features in an image, which
The learned pseudo-words can not only make the model generate high-quality and diverse images, but also can be easily controlled by additional text descriptions. 
%is well compatible with additional descriptions. 
% \pengxu{DreamArtist not only learns concepts in images, but also form, content and context.}
Extensive qualitative and quantitative experimental analyses have demonstrated that our method substantially outperforms existing methods.
Moreover, our DreamArtist method is highly controllable and can be used in combination with complex descriptions, presenting a promising flexibility and potential for deploying other models.
%Furthermore, our method has been able to pass the Turing test, and it has been difficult for humans to distinguish it from highly realistic images.

\begin{figure}[ht!]
    \centering
    \includegraphics[width=0.4\textwidth]{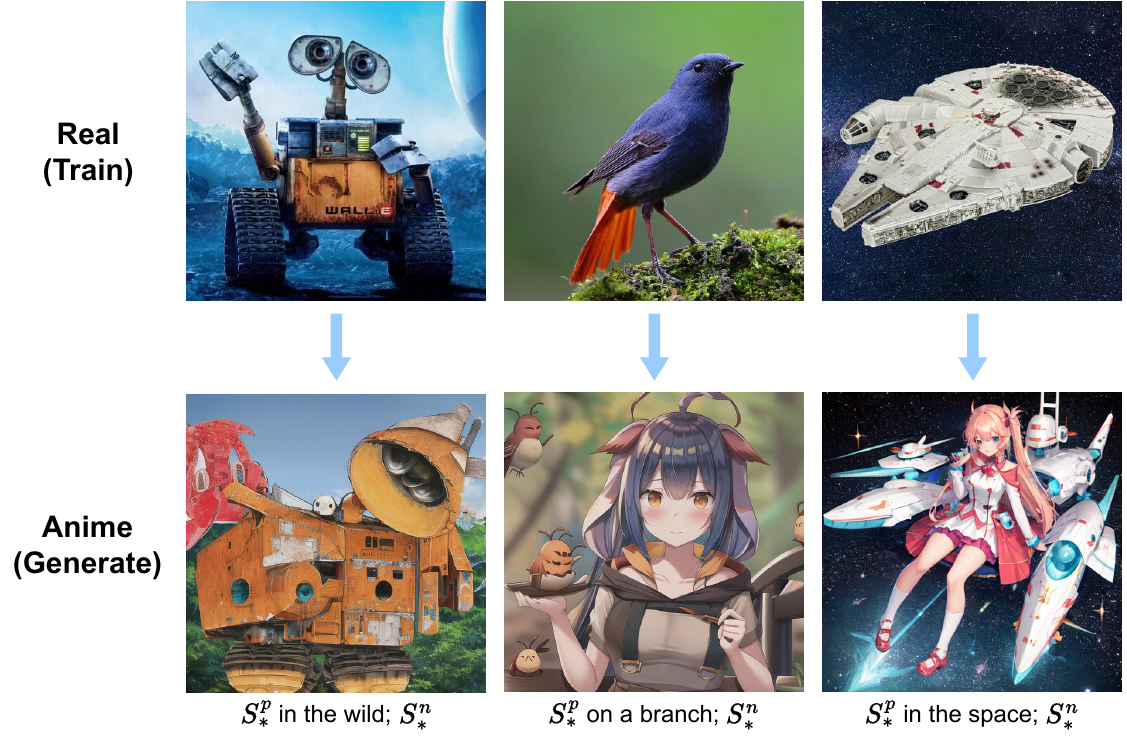}
    %\vspace{-5pt}
    \caption{Limitations of our method with domain shit.}
    \vspace{-10pt}
    \label{fig:exp_domain_shit}
\end{figure}

\begin{figure}[ht!]
    \centering
    \includegraphics[width=0.4\textwidth]{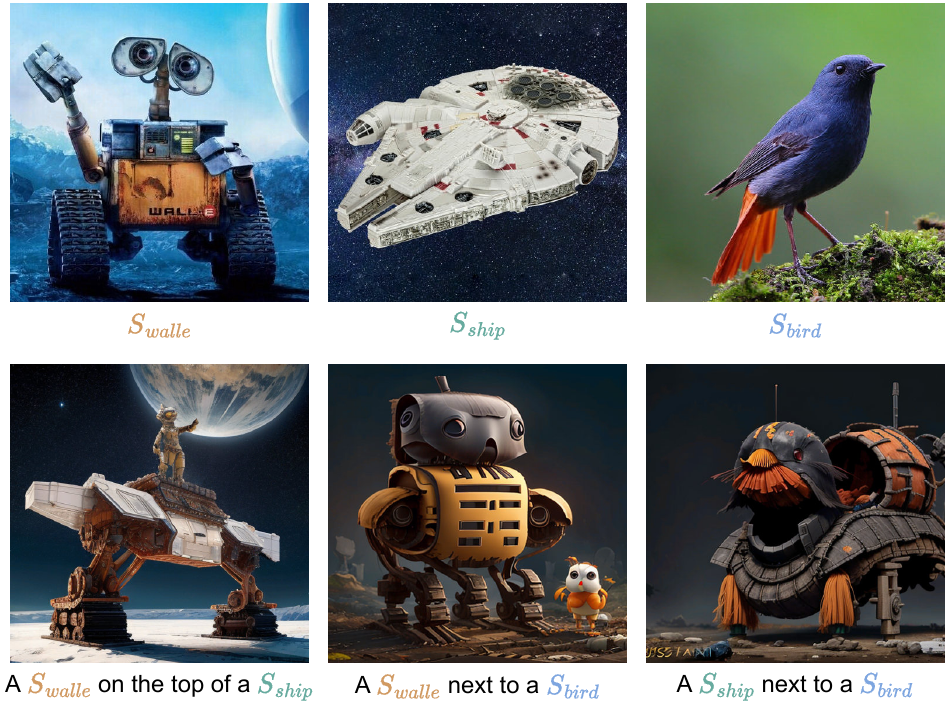}
    %\vspace{-5pt}
    \caption{Limitations of our method in entity concept composition.}
    \vspace{-10pt}
    \label{fig:exp_compose}
\end{figure}

{\Rcolor
\section{Limitations and future work.} 
\noindent\textbf{Domain Shift:}
Our DreamArtist, when trained on a base model within the real-world domain, struggles to accurately render its learned features upon application to a base model in the anime domain, as shown in \cref{fig:exp_domain_shit}. Only partial of the learned features can be effectively rendered, and the generated outputs are heavily influenced by the biases inherent in the anime domain's base model.

\noindent\textbf{Compositions of Entity:}
Our DreamArtist is applicable to concept compositions by combining learned pseudo-words for promising and flexible image generation. However, it sometimes fails on compositions of entity concepts, since they are individually learned from different reference images and may interference with each other. When combining two entity concepts, the model may struggle to accurately render them as distinct entities. As shown in \cref{fig:exp_compose},this often leads to a fusion of their respective features, resulting in mutual interference. The model sometimes fails to distinctly map each concept to its corresponding independent entity.
}

\noindent\textbf{Future Work:}
\revise{Thus, this issue can be solved by continual text-to-image generation with continual learning methods}, to welcome more and more new concepts and ensure highly controllable generation from learned words and original words. 
Besides, domain shift between reference images and original images for pre-training would cause generation failure, which is another issue to address in future work.

\begin{appendices}




\end{appendices}


{\small
\bibliographystyle{bst/sn-aps}
\bibliography{sn-bibliography}

\begin{thebibliography}{10}
\providecommand{\url}[1]{{#1}}
\providecommand{\urlprefix}{URL }
\providecommand{\doi}[1]{\url{https://doi.org/#1}}
\bibcommenthead

\bibitem{ImgGen}
C.~Saharia, W.~Chan, S.~Saxena, L.~Li, J.~Whang, E.~Denton, S.K.S. Ghasemipour, B.K. Ayan, S.S. Mahdavi, R.G. Lopes, T.~Salimans, J.~Ho, D.J. Fleet, M.~Norouzi, Photorealistic text-to-image diffusion models with deep language understanding.
\newblock CoRR \textbf{abs/2205.11487} (2022).
\newblock {\href{https://arxiv.org/abs/2205.11487}{{2205.11487}}}

\bibitem{LDM}
R.~Rombach, A.~Blattmann, D.~Lorenz, P.~Esser, B.~Ommer, \emph{High-Resolution Image Synthesis with Latent Diffusion Models}, in \emph{{IEEE/CVF} Conference on Computer Vision and Pattern Recognition} (2022), pp. 10674--10685

\bibitem{Dalle2}
A.~Ramesh, P.~Dhariwal, A.~Nichol, C.~Chu, M.~Chen, Hierarchical text-conditional image generation with {CLIP} latents.
\newblock CoRR \textbf{abs/2204.06125} (2022).
\newblock {\href{https://arxiv.org/abs/2204.06125}{{2204.06125}}}

\bibitem{sty3}
T.~Karras, M.~Aittala, S.~Laine, E.~H{\"{a}}rk{\"{o}}nen, J.~Hellsten, J.~Lehtinen, T.~Aila, \emph{Alias-Free Generative Adversarial Networks}, in \emph{Advances in Neural Information Processing Systems} (2021), pp. 852--863

\bibitem{DDIM}
J.~Song, C.~Meng, S.~Ermon, \emph{Denoising Diffusion Implicit Models}, in \emph{International Conference on Learning Representations} (2021)

\bibitem{Glow}
D.P. Kingma, P.~Dhariwal, \emph{Glow: Generative Flow with Invertible 1x1 Convolutions}, in \emph{Advances in Neural Information Processing Systems} (2018), pp. 10236--10245

\bibitem{BigGAN}
A.~Brock, J.~Donahue, K.~Simonyan, \emph{Large Scale {GAN} Training for High Fidelity Natural Image Synthesis}, in \emph{International Conference on Learning Representations} (2019)

\bibitem{Palette}
C.~Saharia, W.~Chan, H.~Chang, C.A. Lee, J.~Ho, T.~Salimans, D.J. Fleet, M.~Norouzi, \emph{Palette: Image-to-Image Diffusion Models}, in \emph{{SIGGRAPH} '22: Special Interest Group on Computer Graphics and Interactive Techniques Conference} (2022), pp. 15:1--15:10

\bibitem{DiffGans}
P.~Dhariwal, A.Q. Nichol, \emph{Diffusion Models Beat GANs on Image Synthesis}, in \emph{Advances in Neural Information Processing Systems} (2021), pp. 8780--8794

\bibitem{UNet}
O.~Ronneberger, P.~Fischer, T.~Brox, \emph{U-Net: Convolutional Networks for Biomedical Image Segmentation}, in \emph{Medical Image Computing and Computer-Assisted Intervention, {MICCAI}}, vol. 9351 (2015), pp. 234--241

\bibitem{ScoreDiff}
Y.~Song, J.~Sohl{-}Dickstein, D.P. Kingma, A.~Kumar, S.~Ermon, B.~Poole, \emph{Score-Based Generative Modeling through Stochastic Differential Equations}, in \emph{International Conference on Learning Representations} (2021)

\bibitem{GLIDE}
A.Q. Nichol, P.~Dhariwal, A.~Ramesh, P.~Shyam, P.~Mishkin, B.~McGrew, I.~Sutskever, M.~Chen, \emph{{GLIDE:} Towards Photorealistic Image Generation and Editing with Text-Guided Diffusion Models}, in \emph{International Conference on Machine Learning} (2022), pp. 16784--16804

\bibitem{RiFeGAN}
J.~Cheng, F.~Wu, Y.~Tian, L.~Wang, D.~Tao, \emph{RiFeGAN: Rich Feature Generation for Text-to-Image Synthesis From Prior Knowledge}, in \emph{{IEEE/CVF} Conference on Computer Vision and Pattern Recognition} (2020), pp. 10908--10917

\bibitem{ZeroShotGen}
A.~Jain, B.~Mildenhall, J.T. Barron, P.~Abbeel, B.~Poole, \emph{Zero-Shot Text-Guided Object Generation with Dream Fields}, in \emph{{IEEE/CVF} Conference on Computer Vision and Pattern Recognition} (2022), pp. 857--866

\bibitem{ctrlGAN}
B.~Li, X.~Qi, T.~Lukasiewicz, P.H.S. Torr, \emph{Controllable Text-to-Image Generation}, in \emph{Advances in Neural Information Processing Systems} (2019), pp. 2063--2073

\bibitem{MirrorGAN}
T.~Qiao, J.~Zhang, D.~Xu, D.~Tao, \emph{MirrorGAN: Learning Text-To-Image Generation by Redescription}, in \emph{{IEEE} Conference on Computer Vision and Pattern Recognition} (2019), pp. 1505--1514

\bibitem{DFGAN}
M.~Tao, H.~Tang, S.~Wu, N.~Sebe, F.~Wu, X.~Jing, {DF-GAN:} deep fusion generative adversarial networks for text-to-image synthesis.
\newblock CoRR \textbf{abs/2008.05865} (2020).
\newblock {\href{https://arxiv.org/abs/2008.05865}{{2008.05865}}}

\bibitem{TIGAN}
S.E. Reed, Z.~Akata, X.~Yan, L.~Logeswaran, B.~Schiele, H.~Lee, \emph{Generative Adversarial Text to Image Synthesis}, in \emph{Proceedings of the International Conference on Machine Learning}, vol.~48 (2016), pp. 1060--1069

\bibitem{VQGAN-CLIP}
K.~Crowson, S.~Biderman, D.~Kornis, D.~Stander, E.~Hallahan, L.~Castricato, E.~Raff, \emph{{VQGAN-CLIP:} Open Domain Image Generation and Editing with Natural Language Guidance}, in \emph{ECCV} (2022), pp. 88--105

\bibitem{PPGAN}
Y.~Liu, J.~Peng, J.J.Q. Yu, Y.~Wu, \emph{{PPGAN:} Privacy-Preserving Generative Adversarial Network}, in \emph{{IEEE} International Conference on Parallel and Distributed Systems} (2019), pp. 985--989

\bibitem{guid_cls}
X.~Liu, D.H. Park, S.~Azadi, G.~Zhang, A.~Chopikyan, Y.~Hu, H.~Shi, A.~Rohrbach, T.~Darrell, More control for free! image synthesis with semantic diffusion guidance.
\newblock CoRR \textbf{abs/2112.05744} (2021).
\newblock {\href{https://arxiv.org/abs/2112.05744}{{2112.05744}}}

\bibitem{guid_cls_free}
J.~Ho, T.~Salimans, \emph{Classifier-Free Diffusion Guidance}, in \emph{NeurIPS 2021 Workshop on Deep Generative Models and Downstream Applications} (2021)

\bibitem{CogView}
M.~Ding, Z.~Yang, W.~Hong, W.~Zheng, C.~Zhou, D.~Yin, J.~Lin, X.~Zou, Z.~Shao, H.~Yang, J.~Tang, \emph{CogView: Mastering Text-to-Image Generation via Transformers}, in \emph{Advances in Neural Information Processing Systems} (2021), pp. 19822--19835

\bibitem{Parti}
J.~Yu, Y.~Xu, J.Y. Koh, T.~Luong, G.~Baid, Z.~Wang, V.~Vasudevan, A.~Ku, Y.~Yang, B.K. Ayan, B.~Hutchinson, W.~Han, Z.~Parekh, X.~Li, H.~Zhang, J.~Baldridge, Y.~Wu, Scaling autoregressive models for content-rich text-to-image generation.
\newblock CoRR \textbf{abs/2206.10789} (2022).
\newblock {\href{https://arxiv.org/abs/2206.10789}{{2206.10789}}}

\bibitem{ERNIE}
H.~Zhang, W.~Yin, Y.~Fang, L.~Li, B.~Duan, Z.~Wu, Y.~Sun, H.~Tian, H.~Wu, H.~Wang, Ernie-vilg: Unified generative pre-training for bidirectional vision-language generation.
\newblock CoRR \textbf{abs/2112.15283} (2021).
\newblock {\href{https://arxiv.org/abs/2112.15283}{{2112.15283}}}

\bibitem{DAAM}
R.~Tang, A.~Pandey, Z.~Jiang, G.~Yang, K.~Kumar, J.~Lin, F.~Ture, What the {DAAM:} interpreting stable diffusion using cross attention.
\newblock CoRR \textbf{abs/2210.04885} (2022).
\newblock {\href{https://arxiv.org/abs/2210.04885}{{2210.04885}}}

\bibitem{TI}
R.~Gal, Y.~Alaluf, Y.~Atzmon, O.~Patashnik, A.H. Bermano, G.~Chechik, D.~Cohen{-}Or, \emph{An Image is Worth One Word: Personalizing Text-to-Image Generation using Textual Inversion}, in \emph{International Conference on Learning Representations} (2023)

\bibitem{DreamBooth}
N.~Ruiz, Y.~Li, V.~Jampani, Y.~Pritch, M.~Rubinstein, K.~Aberman, Dreambooth: Fine tuning text-to-image diffusion models for subject-driven generation.
\newblock CoRR \textbf{abs/2208.12242} (2022).
\newblock {\href{https://arxiv.org/abs/2208.12242}{{2208.12242}}}

\bibitem{LoRA}
E.J. Hu, Y.~Shen, P.~Wallis, Z.~Allen{-}Zhu, Y.~Li, S.~Wang, L.~Wang, W.~Chen, \emph{LoRA: Low-Rank Adaptation of Large Language Models}, in \emph{{ICLR} The Tenth International Conference on Learning Representations} (2022)

\bibitem{LAION}
C.~Schuhmann, R.~Beaumont, R.~Vencu, C.~Gordon, R.~Wightman, M.~Cherti, T.~Coombes, A.~Katta, C.~Mullis, M.~Wortsman, P.~Schramowski, S.~Kundurthy, K.~Crowson, L.~Schmidt, R.~Kaczmarczyk, J.~Jitsev, {LAION-5B:} an open large-scale dataset for training next generation image-text models.
\newblock CoRR \textbf{abs/2210.08402} (2022).
\newblock {\href{https://arxiv.org/abs/2210.08402}{{2210.08402}}}

\bibitem{danbooru2021}
Anonymous, D.~community, G.~Branwen.
\newblock Danbooru2021: A large-scale crowdsourced and tagged anime illustration dataset (2022).
\newblock \urlprefix\url{https://www.gwern.net/Danbooru2021}

\bibitem{Cnet}
L.~Zhang, A.~Rao, M.~Agrawala, \emph{Adding Conditional Control to Text-to-Image Diffusion Models}, in \emph{{IEEE/CVF} International Conference on Computer Vision} (2023), pp. 3813--3824

\bibitem{T2IA}
C.~Mou, X.~Wang, L.~Xie, Y.~Wu, J.~Zhang, Z.~Qi, Y.~Shan, \emph{T2I-Adapter: Learning Adapters to Dig Out More Controllable Ability for Text-to-Image Diffusion Models}, in \emph{Thirty-Eighth {AAAI} Conference on Artificial Intelligence, {AAAI} 2024} (2024), pp. 4296--4304

\bibitem{IPA}
H.~Ye, J.~Zhang, S.~Liu, X.~Han, W.~Yang, Ip-adapter: Text compatible image prompt adapter for text-to-image diffusion models.
\newblock CoRR \textbf{abs/2308.06721} (2023).
\newblock {\href{https://arxiv.org/abs/2308.06721}{{2308.06721}}}

\bibitem{customdiffusion}
N.~Kumari, B.~Zhang, R.~Zhang, E.~Shechtman, J.Y. Zhu, \emph{Multi-Concept Customization of Text-to-Image Diffusion}, in \emph{{IEEE} Conference on Computer Vision and Pattern Recognition} (2023)

\bibitem{DDPM}
J.~Ho, A.~Jain, P.~Abbeel, \emph{Denoising Diffusion Probabilistic Models}, in \emph{Advances in Neural Information Processing Systems} (2020)

\bibitem{PTuning2}
X.~Liu, K.~Ji, Y.~Fu, Z.~Du, Z.~Yang, J.~Tang, P-tuning v2: Prompt tuning can be comparable to fine-tuning universally across scales and tasks.
\newblock CoRR \textbf{abs/2110.07602} (2021).
\newblock {\href{https://arxiv.org/abs/2110.07602}{{2110.07602}}}

\bibitem{PromptZeroshot}
T.B. Brown, B.~Mann, N.~Ryder, M.~Subbiah, J.~Kaplan, P.~Dhariwal, A.~Neelakantan, P.~Shyam, G.~Sastry, A.~Askell, S.~Agarwal, A.~Herbert{-}Voss, G.~Krueger, T.~Henighan, R.~Child, A.~Ramesh, D.M. Ziegler, J.~Wu, C.~Winter, C.~Hesse, M.~Chen, E.~Sigler, M.~Litwin, S.~Gray, B.~Chess, J.~Clark, C.~Berner, S.~McCandlish, A.~Radford, I.~Sutskever, D.~Amodei, \emph{Language Models are Few-Shot Learners}, in \emph{Advances in Neural Information Processing Systems} (2020)

\bibitem{PromptCloze}
T.~Schick, H.~Sch{\"{u}}tze, \emph{Exploiting Cloze-Questions for Few-Shot Text Classification and Natural Language Inference}, in \emph{Proceedings of the 16th Conference of the European Chapter of the Association for Computational Linguistics, {EACL}} (2021), pp. 255--269

\bibitem{CoPo}
K.~Zhou, J.~Yang, C.C. Loy, Z.~Liu, Learning to prompt for vision-language models.
\newblock Int. J. Comput. Vis. \textbf{130}(9), 2337--2348 (2022)

\bibitem{CoCoPo}
K.~Zhou, J.~Yang, C.C. Loy, Z.~Liu, \emph{Conditional Prompt Learning for Vision-Language Models}, in \emph{{IEEE/CVF} Conference on Computer Vision and Pattern Recognition} (2022), pp. 16795--16804

\bibitem{PTVision}
K.~Ding, Y.~Wang, P.~Liu, Q.~Yu, H.~Zhang, S.~Xiang, C.~Pan, Prompt tuning with soft context sharing for vision-language models.
\newblock CoRR \textbf{abs/2208.13474} (2022).
\newblock {\href{https://arxiv.org/abs/2208.13474}{{2208.13474}}}

\bibitem{PromptTuning}
B.~Lester, R.~Al{-}Rfou, N.~Constant, \emph{The Power of Scale for Parameter-Efficient Prompt Tuning}, in \emph{Proceedings of the Conference on Empirical Methods in Natural Language Processing, {EMNLP}} (2021), pp. 3045--3059

\bibitem{PrefixTuning}
X.L. Li, P.~Liang, \emph{Prefix-Tuning: Optimizing Continuous Prompts for Generation}, in \emph{Proceedings of the 59th Annual Meeting of the Association for Computational Linguistics and the 11th International Joint Conference on Natural Language Processing, {ACL/IJCNLP}} (2021), pp. 4582--4597

\bibitem{LPIPS}
R.~Zhang, P.~Isola, A.A. Efros, E.~Shechtman, O.~Wang, \emph{The Unreasonable Effectiveness of Deep Features as a Perceptual Metric}, in \emph{{IEEE} Conference on Computer Vision and Pattern Recognition} (2018), pp. 586--595

\bibitem{styloss}
L.A. Gatys, A.S. Ecker, M.~Bethge, \emph{Image Style Transfer Using Convolutional Neural Networks}, in \emph{{IEEE} Conference on Computer Vision and Pattern Recognition} (2016), pp. 2414--2423

\end{thebibliography}
}


\end{document}